%% file: main.tex
\documentclass[runningheads]{llncs}

\usepackage{eccv}

\usepackage{eccvabbrv}

\usepackage{graphicx}
\usepackage{booktabs}

\usepackage[accsupp]{axessibility}  

\usepackage[pagebackref,breaklinks,colorlinks,citecolor=eccvblue]{hyperref}

\usepackage{orcidlink}
\usepackage{array}
\usepackage{multirow}
\usepackage[numbers]{natbib}
\usepackage{multicol}
\usepackage{graphicx}
\usepackage{amsmath}
\usepackage{algorithm}
\usepackage{wrapfig}
\usepackage{algorithmic}
\usepackage{caption}
\newtheorem{innercustomthm}{Theorem}
\newenvironment{customthm}[1]
  {\renewcommand\theinnercustomthm{#1}\innercustomthm}
  {\endinnercustomthm}

\usepackage{titlesec}
\titleclass{\part}{top}
\titleformat{\part}[display]
  {\normalfont\huge\bfseries}{\centering\partname\ \thepart}{20pt}{\huge\centering}

\makeatletter
\newcommand\thankssymb[1]{\textsuperscript{\@fnsymbol{#1}}}
\makeatother

\usepackage[toc,page,header]{appendix}
\usepackage{minitoc}
\noptcrule
\renewcommand \thepart{}
\renewcommand \partname{}

\begin{document}

\title{\includegraphics[height=1em]{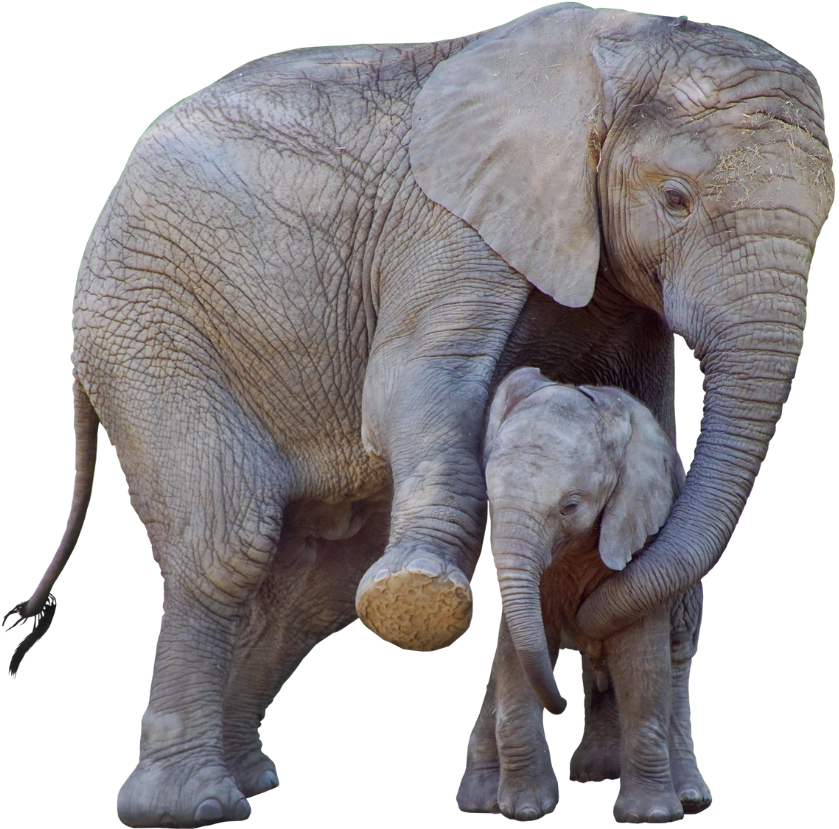} MAMA: Meta-optimized Angular Margin Contrastive Framework for Video-Language Representation Learning}

\titlerunning{MAMA: Meta-optimized Angular MArgin Contrastive Framework}

\author{Thong Nguyen\inst{1}\thanks{First author} \and
Yi Bin\inst{1}\thanks{Corresponding author} \and
Xiaobao Wu\inst{2} \and 
Xinshuai Dong\inst{3} \and
Zhiyuan Hu\inst{1} \and \\
Khoi Le\inst{4} \and 
Cong-Duy Nguyen\inst{2} \and
See-Kiong Ng\inst{1} \and
Luu Anh Tuan\inst{2}}


\institute{National University of Singapore (NUS), Singapore \and
Nanyang Technological University (NTU), Singapore \and
Carnegie Mellon University (CMU), USA \and
VinAI Research, Vietnam\\
 \href{https://nguyentthong.github.io/MAMA}{nguyentthong.github.io/MAMA}} 
\maketitle

\vspace{-20pt}
\begin{abstract}

Data quality stands at the forefront of deciding the effectiveness of video-language representation learning. However, video-text pairs in previous data typically do not align perfectly with each other, which might lead to video-language representations that do not accurately reflect cross-modal semantics. Moreover, previous data also possess an uneven distribution of concepts, thereby hampering the downstream performance across unpopular subjects. To address these problems, we propose \textbf{MAMA}, a new approach to learning video-language representations by utilizing a contrastive objective with a subtractive angular margin to regularize cross-modal representations in their effort to reach perfect similarity. Furthermore, to adapt to the non-uniform concept distribution, MAMA utilizes a multi-layer perceptron (MLP)-parameterized weighting function that maps loss values to sample weights which enable dynamic adjustment of the model’s focus throughout the training. With the training guided by a small amount of unbiased meta-data and augmented by video-text data generated by large vision-language model, MAMA improves video-language representations and achieve superior performances on commonly used video question answering and text-video retrieval datasets. The code, model, and data have been made available at \href{https://nguyentthong.github.io/MAMA}{nguyentthong.github.io/MAMA}.


\end{abstract}

\input{files/01_introduction}
\input{files/02_related_work}
\input{files/03_methodology}
\input{files/04_experiments}
\input{files/05_conclusion}
\input{files/06_acknowledgement}

%
%
\bibliographystyle{splncs04}
\bibliography{main}


\appendix
\input{files/appendix}

\end{document}

%% file: files/01_introduction.tex
\section{Introduction}

\begin{figure*}[t]
    \centering
    \caption{Examples of video inputs and their textual descriptions.}
    \label{fig:example_figure}
    \begin{subfigure}[h!]{0.48\linewidth}
    \includegraphics[width=\linewidth]{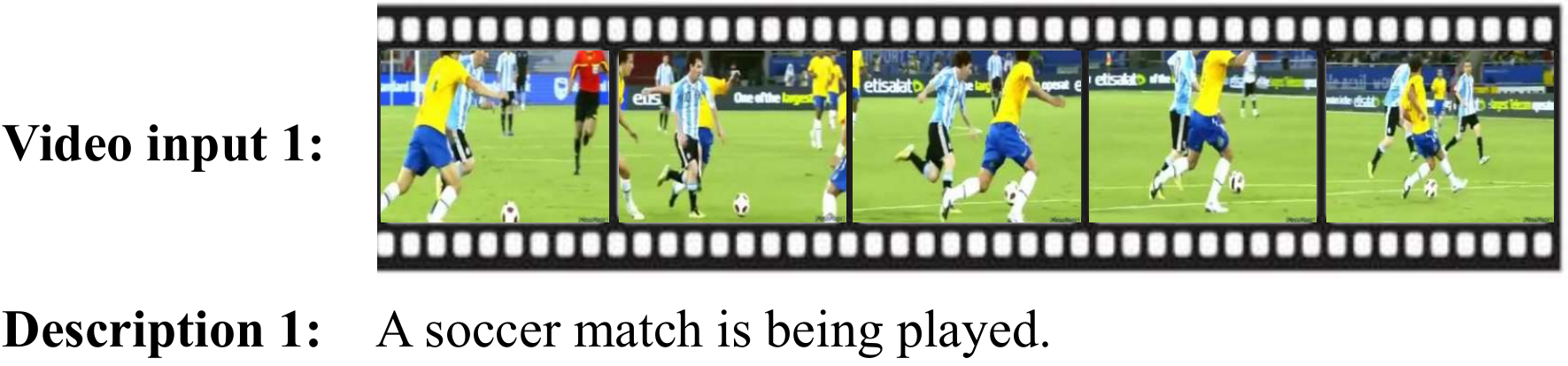}
    \end{subfigure}
    \quad
    \begin{subfigure}[h!]{0.48\linewidth}
    \includegraphics[width=\linewidth]{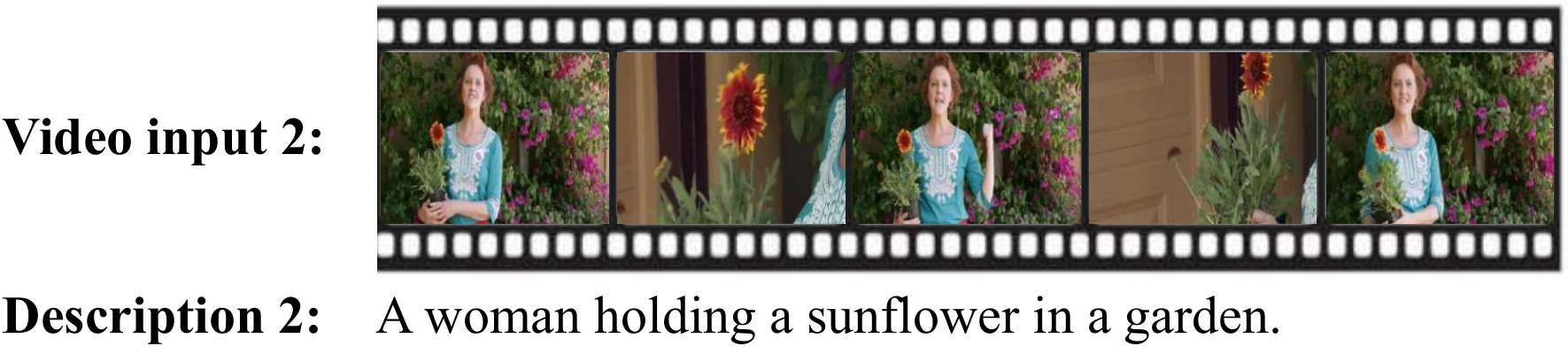}
    \end{subfigure}
    \vspace{-2mm}
\end{figure*}

\begin{figure*}[t]
    \centering
    \caption{Topic distribution of the MSRVTT dataset. We use Latent Dirichlet Allocation (LDA) to extract topics from manually annotated descriptions of videos.}
    \label{fig:distribution_figure}
    \includegraphics[width=0.5\linewidth]{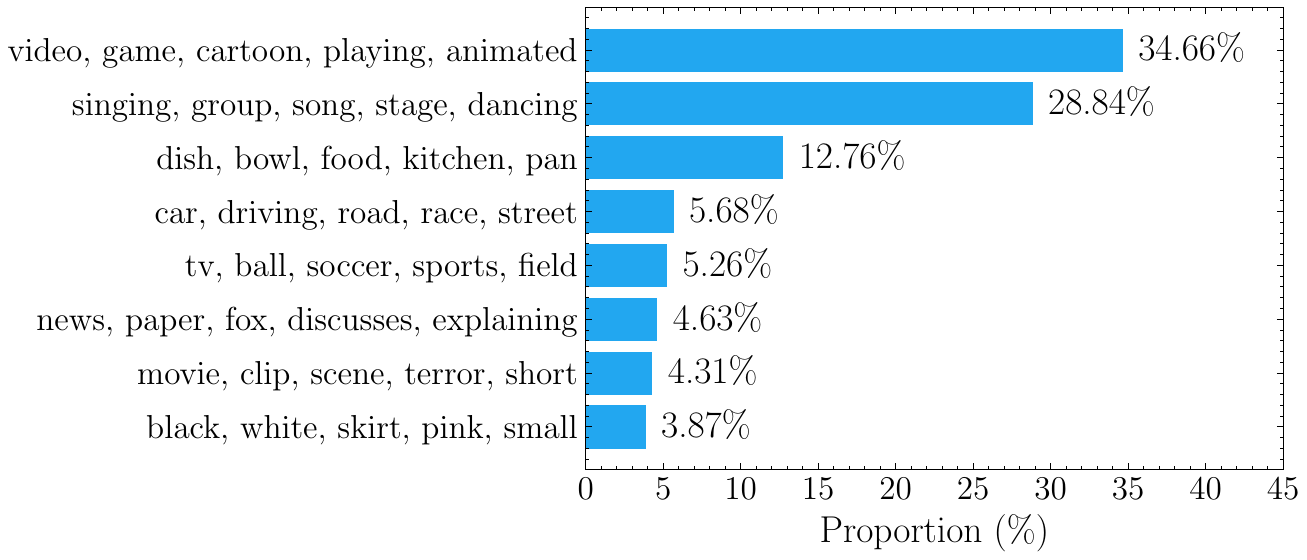}
    \vspace{-8mm}
\end{figure*}

Learning vision and language representations has advanced state-of-the-art across multiple cross-modal tasks, such as video question answering (VideoQA) \citep{lei2021less, li2020hero, fu2021violet, fu2023empirical, cheng2023vindlu, seo2022end, wang2023all} and text-video retrieval \citep{zhu2020actbert, tang2021decembert, xu2021vlm, lei2022revealing, li2023lavender, wang2022omnivl, buch2022revisiting}. The success of vision-language representation learning mainly results from the effectiveness of contrastive learning, which projects video and text inputs into a common latent space based on their semantic similarities.

Data quality stands at the forefront of influencing the efficacy of video-language representation learning, particularly through its cleanliness and diversity, which are pivotal in optimizing model performance and bolstering generalization capabilities \citep{zhao2023learning}. Within this context, we pinpoint two interrelated issues that significantly impact the \textit{cleanliness} and \textit{diversity} of video-text data: (1) the imperfection in alignment, characterized by a scarcity of fine-grained details, and (2) the imbalance among the concepts of data samples. First, a video and its textual description commonly does not perfectly align with each other, \textit{e.g.} the description may omit certain details in the video, such as the green grass on a soccer field or the pink flowers at the background in two videos of Fig. \ref{fig:example_figure}. As such, aggressively minimizing a contrastive loss to pull video and language representations together might result in distorted video-language representations that do not closely capture the semantic similarity of video-text pairs, thus compromising the interpretive video-language understanding. Second, as an example, Fig. \ref{fig:distribution_figure} shows a majority of video-text pairs in the well-known MSRVTT dataset denote the video game or singing topic (the topic bars), while a minority of them describe the fashion topic (the bottom bar). This might skew the model’s exposure towards certain topics at the expense of others, undermining the model’s ability to perform uniformly across a broad spectrum of subjects \citep{li2022trustworthy}. Addressing these intertwined challenges is crucial for advancing video-language representation learning and facilitating more effective video-language understanding.

Our method, called \textbf{MAMA}: \textbf{M}eta-optimized \textbf{A}ngular \textbf{MA}rgin Contrastive Learning, subtracts a margin between a positive video and text sample in the angular space. Our mathematical derivation shows that the subtracted margin can decay the gradient norm, thus providing a regularization effect to constrain positive but imperfectly aligned samples from reaching perfect similarity. For the imbalance issue, to enable the network to dynamically adjust its focus during training, MAMA introduces a sample reweighting strategy that maps loss values to sample weights. A natural idea is to assign higher weights to larger losses to emphasize minority classes for which the network is likely to make mistakes \citep{sun2007cost, malisiewicz2011ensemble}. However, with respect to the imperfect alignment issue, it is advisable to assign higher weights to smaller losses, as there is a greater chance that these samples better align with each other, hence forming cleaner samples which the network should focus on \cite{zhang2018generalized, wang2017robust}. To avoid the exhaustive effort in manually specifying a weighting function and enhance the function generalization, we parameterize our weighting function as a multi-layer perceptron (MLP) as theoretically a universal approximator for any continuous function \citep{csaji2001approximation}, and use a small unbiased validation set (meta-data) to train the MLP.

Moreover, to further enhance the diversity in video-language representation learning, MAMA utilizes off-the-shelf large vision-language model (LVLM) to augment downstream video-text data. In particular, given an additional video input, we adopt the density peak-based clustering approach to extract its key frames, then concatenate the extracted frames into one grid image, and forward the image with a relevant input prompt to obtain the text pairing. Combined with the LVLM-augmented video-text data, MAMA considerably outperforms previous state-of-the-art video-language representation learning methods on standard MSRVTT, DiDeMo, ActivityNet, TGIF-QA-R, NExT-QA, and Causal-VidQA datasets.

%% file: files/02_related_work.tex
\vspace{-4mm}
\section{Related Work}
\vspace{-2mm}
\textbf{Video-language representation learning.} Recent approaches utilizing contrastive learning have demonstrated impressive performance in video-language understanding tasks including videoQA \citep{wang2019holistic,cheng2023vindlu, zellers2021merlot, lei2022revealing, wang2022omnivl, buch2022revisiting} and text$\leftrightarrow$video retrieval \citep{akbari2021vatt, luo2022clip4clip, lei2021understanding, sun2019learning, luo2020univl, zhao2023learning, xu2021videoclip, bain2021frozen, lin2022eclipse, gao2021clip2tv, bain2022clip, xue2022clip, nguyen2024read, nguyen2024video}. VideoQA requires the model to infer an answer based on the fusion of video and question representations, while text$\leftrightarrow$video retrieval requires a model to project text query and video to a common latent space, where their similarity is directly calculated. Collecting video-text data is expensive and time-consuming. Therefore, previous methods take advantage of audio transcripts \citep{yang2023vid2seq, yu2021vision, luo2020univl,wei2023multi,wei2022audio} or adapt image-text models to videos \citep{lei2022revealing, buch2022revisiting, lei2021less,wei2024learning}. Unfortunately, audio transcripts often do not temporally align with the video content \citep{han2022temporal} and image-text models are ineffective in temporal reasoning \citep{lei2022revealing}. Instead,  Zhao et al. \citep{zhao2023learning} and 
Yang et al. \citep{yang2023vid2seq} fine-tune an LLM to be conditioned on visual frames to augment textual descriptions for videos. However, due to the huge size of LLMs, their approaches incur huge computational and storage cost. 

\noindent\textbf{Deep metric learning (DML).} Contrastive learning is an implementation of DML. Its target is to learn a function that maps data samples into a latent space, where semantically similar objects stay close while dissimilar ones are far away \citep{sohn2016improved, ma2021contrastive, nguyen2023demaformer, nguyen2021contrastive, nguyen2022adaptive, nguyen2023improving, nguyen2024kdmcse, nguyen2024topic,wu2020short,wu2022mitigating,wu2023infoctm,wu2024dynamic}. Numerous efforts introduce angular margin between two input samples to enhance discriminative power of input representations \citep{deng2019arcface, coria2020metric}. Nevertheless, little attention is paid to the imperfect alignment between two positive samples. The situation is more challenging for video and language, since these modalities exhibit a more distinct semantic gap than any cross-modal input pairs. 

\noindent\textbf{Sample Reweighting.} To manage the impact of each sample upon the training process, sample reweighting pre-evaluates samples by mapping training loss to sample weight, and dynamically modulating the weight throughout training. There are two approaches to design the mapping function. The first approach monotonically increases the weight when the loss increases, which adapts to the class imbalance case as minor classes tend to engender higher loss values. Methods that belong to this first group include focal loss \cite{bin2023non,lin2018focal}, hard example mining \cite{li2023your,malisiewicz2011ensemble}, and boosting algorithm \citep{freund1997decision}. The second approach monotonically decreases the weight for larger training loss values \citep{jiang2014self, fernando2003reweight,zhang2018generalized}. In particular, self-paced learning (SPL) \citep{jiang2014self} and iterative reweighting \cite{fernando2003reweight, zhang2018generalized} put more emphasis upon cleaner samples with smaller losses. Both approaches need to pre-specify the scheme of their weighting function. This could restrain their flexibility to adapt to diverse scenarios in practice, \textit{e.g.} training data might possess both label noise and and class imbalance issue.

%% file: files/03_methodology.tex
\begin{figure*}[t]
    \centering
\includegraphics[width=\linewidth]{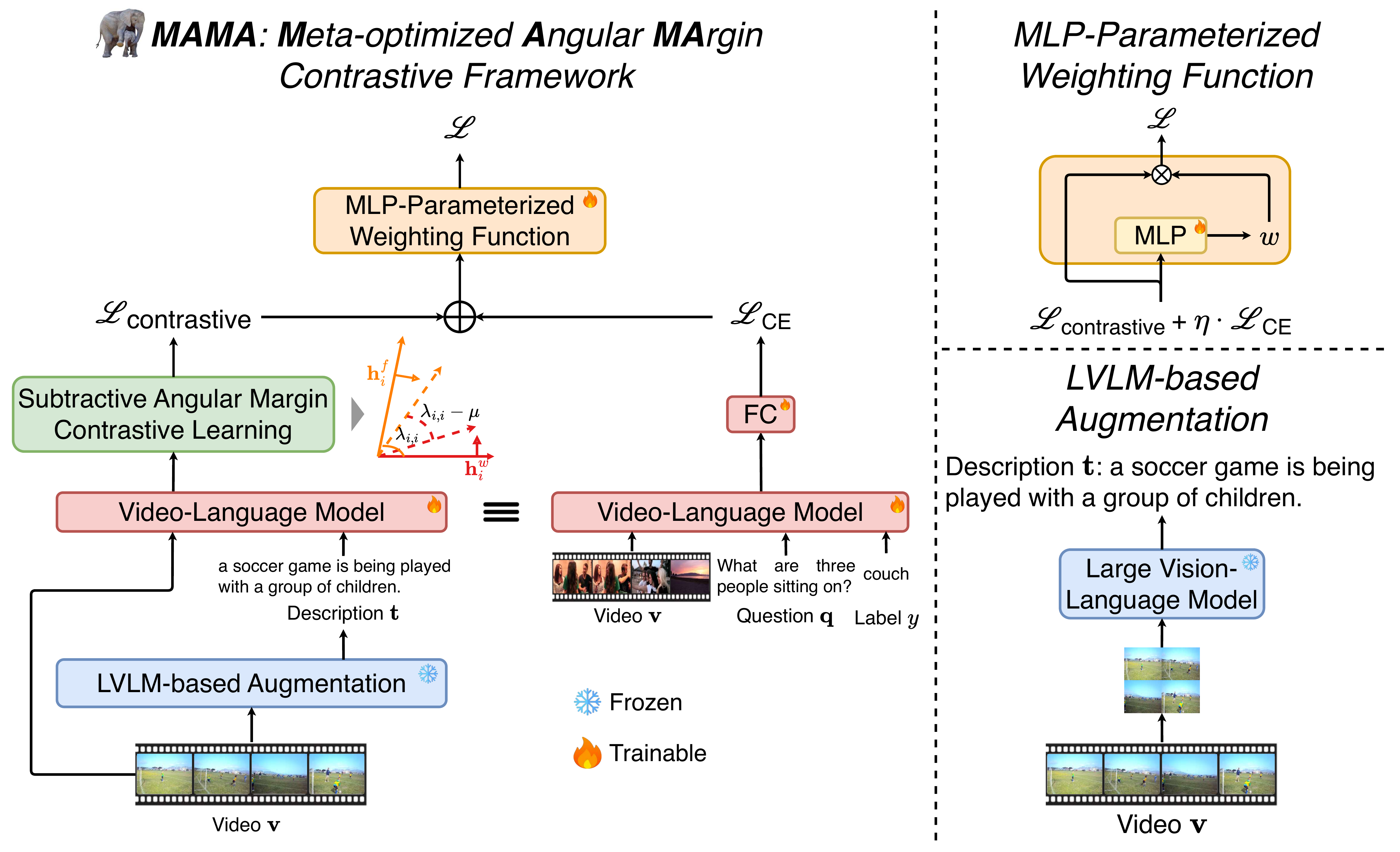}
    \caption{Illustration of the proposed MAMA framework and its components.} 
    \label{fig:overall_illustration}
    \vspace{-5mm}
\end{figure*}
\vspace{-2mm}
\section{MAMA Framework}
\vspace{-2mm}
In this section, we explain MAMA, the meta-optimized angular margin contrastive framework with our augmentation strategy for video-language representation learning. We provide an overall illustration of MAMA in Figure \ref{fig:overall_illustration}.
\vspace{-2mm}
\subsection{Video-Language Representation Learning}
\label{sect:video_language_representation_learning}
We are given a corpus of video data, in which each video $\mathbf{v}_{i}$ is attached with a textual description $\mathbf{t}_{i}$ and possibly a label $y_{i}$. In the beginning, we embed the video input $\mathbf{v}_{i}$ into a sequence of visual representations $\mathbf{V}_{i}^{f} = \{\mathbf{h}_{i,j}^{f}\}_{j=1}^{N_{\mathbf{v}_{i}}}$, where $N_{\mathbf{v}_{i}}$ is the number of randomly sampled video frames in $\mathbf{v}_{i}$. We also embed the textual description into a sequence of representations $\mathbf{T}^{w}_{i} = \{\mathbf{h}_{i,j}^{w}\}_{j=1}^{N_{\mathbf{t}_{i}}}$, where $N_{\mathbf{t}_{i}}$ is the number of words in $\mathbf{t}_{i}$.

In this work, we conduct video-language representation learning for two types of video-language model, \textit{i.e.} dual and bidirectional model. For the dual type, we will pool the visual representations $\mathbf{V}^{f}_{i}$ and textual representations $\mathbf{T}^{w}_{i}$ to obtain global representations $\mathbf{h}^{V}_{i}$ and $\mathbf{h}^{T}_{i}$ for the video and text input, respectively. We use cosine similarity to calculate the cross-modal similarity $S_{\mathbf{v}_{i}, \mathbf{t}_{i}}$. Then, we aim to maximize the similarities $S_{\mathbf{v}_{i}, \mathbf{t}_{i}}$ of video-text pairs within the data, relative to those of the unaligned video-text pairs.

The bidirectional model will concatenate the visual and textual representations into a sequence, then feed the sequence to Transformer attention layers to capture the video-text relations. Subsequently, we will forward the [CLS] token which represents the video-text sequence to a fully-connected (FC) layer. The FC layer will calculate the similarity $S_{\mathbf{v}_{i}, \mathbf{t}_{i}}$ if the video input $\mathbf{v}_{i}$ has a text pairing $\mathbf{t}_{i}$, or calculate the log-likelihood $\log q(y_{i} | \mathbf{v}_{i}, \mathbf{t}_{i})$ of the answer if the video input $\mathbf{v}_{i}$ has a label $y_{i}$. In the former case, similar to the dual type we also maximize the similarity of in-distribution video-text pairs, whereas for the latter case we maximize the log-likelihood of the answer label $y_{i}$.

In general, for all video inputs, we maximize the similarity $S_{\mathbf{v}, \mathbf{t}}$ through minimizing the cross-modal contrastive loss as:
\begin{equation}
\hspace{-35pt}
\small
\mathcal{L}^{v,t}_{\text{contrastive}, i} = -\log \frac{e^{S_{\textbf{v}_{i}, \textbf{t}_{i}}/\tau}}{\sum\limits_{j=1}^{B}e^{S_{\textbf{v}_{i}, \textbf{t}_{j}}/\tau}}, \;\;\;\; \mathcal{L}^{t,v}_{\text{contrastive}, i} = - \log \frac{e^{S_{\textbf{t}_{i}, \textbf{v}_{i}}/\tau}}{\sum\limits_{j=1}^{B}e^{S_{\textbf{t}_{i}, \textbf{v}_{j}}}/\tau},
\end{equation}
where $B$ is the batch size and $\tau$ is the temperature hyperparameter. If a video $\mathbf{v}_{i}$ additionally exhibits an answer label $y_{i}$, we jointly combine the cross-entropy loss to maximize the log-likelihood of the label with the above contrastive loss:
\vspace{-4mm}
\begin{gather}
\small
\mathcal{L}_{\text{CE}, i} = - \sum\limits_{y_{i}} p(y_{i}|\mathbf{v}_{i}, \mathbf{t}_{i}) \log q(y_{i}|\mathbf{v}_{i}, \mathbf{t}_{i}), \\  \mathcal{L}_{\text{train},i} =  \mathcal{L}^{v,t}_{\text{contrastive},i} + \mathcal{L}^{t,v}_{\text{contrastive},i} + \eta \cdot \mathcal{L}_{\text{CE},i},
\label{eq:total_training_objective}
\end{gather}
\noindent where $\eta$ is a hyperparameter to balance both losses. It is worth noting that $\eta = 0$ if a video input $\mathbf{v}_{i}$ is not attached with $y_{i}$.
\vspace{-10pt}
\subsection{Meta-optimized Angular Margin Contrastive Framework}
\setlength{\textfloatsep}{-0.1mm}
\begin{algorithm}[t]
\footnotesize	
    \renewcommand{\algorithmicrequire}{\textbf{Input:}}
    \renewcommand{\algorithmicensure}{\textbf{Output:}}
    \caption{Our meta-optimized learning framework}
    \label{alg:example}
    \begin{algorithmic}[1]  \small
        \REQUIRE  Training data $\mathcal{D}$, meta-data $\widehat{\mathcal{D}}$, training batch size $B$, meta-data batch size $M$, initialized video-language model parameter $\Theta^{(0)}$ and MLP network parameter $\theta^{(0)}$
        \ENSURE  Video-language model parameter $\Theta^{(K)}$
        \FOR{$k=0$ {\bfseries to} $K-1$}
        \STATE Sample a training minibatch $\{\textbf{v},\textbf{t},y\}$.
        \STATE Sample a meta-data minibatch $\{\textbf{v}^{\text{meta}},\textbf{t}^{\text{meta}},y^{\text{meta}}\}$.
        \STATE Estimate $\hat{\Theta}^{(t)}$ by Eq. (\ref{eq:estimate_video_language_model}).
        \STATE Update $\theta^{(k+1)}$ by Eq. (\ref{eq:update_mlp_network}).
        \STATE Update $\Theta^{(k+1)}$ by Eq. (\ref{eq:update_video_language_model}).
        \ENDFOR
    \end{algorithmic}
    \label{algo:meta_optimized_framework}
\end{algorithm}
\noindent\textbf{Subtractive angular margin contrastive learning.} 
As cosine similarity is used to calculate $S_{\mathbf{v}_{i}, \mathbf{t}_{j}} \in [-1, 1]$, we denote the angle between the representation of video $i$ and text $j$ as:
\begin{equation}
\small
\lambda_{i,j} = \arccos \left(S_{\mathbf{v}_{i}, \mathbf{t}_{j}}\right).
\label{eq:subtractive_angular_margin_contrastive}
\end{equation}
The original video-language representation learning minimizes $\lambda_{i,i}$ to approach a 0-degree angle. However, since the video and its textual description commonly does not perfectly align with each other, we want the gradient to be regularized when the similarity of positive pair $\textbf{v}_{i}$ and $\textbf{t}_{i}$ becomes small. Particularly, we replace $\mathcal{L}_{\text{contrastive},i}^{v,t}$ by a new training objective:
\[
\mathcal{L}_{\text{angular},i}^{v,t} =
\begin{cases}  
-\log\frac{e^{\cos\left([\lambda_{i,i}-\mu]_{+}\right)/\tau}}{e^{\cos\left([\lambda_{i,i}-\mu]_{+}\right)/\tau} + \sum\limits_{j \neq i} e^{\cos(\lambda_{i,j})/\tau}}, & \text{if } \lambda_{i,i} \leq \frac{\pi}{2} \\
-\log\frac{e^{\cos(\lambda_{i,i})/\tau}}{e^{\cos(\lambda_{i,i})/\tau} + \sum\limits_{j \neq i} e^{\cos(\lambda_{i,j})/\tau}}, & \text{otherwise},
\end{cases}
\]

\noindent A similar objective $\mathcal{L}_{\text{angular},i}^{t,v}$ is formulated between $\mathbf{t}_i$ and $\mathbf{v}_i$. As long as the angular difference $\lambda_{i,i}$ is smaller than $\mu$, the similarity score will become 1. On the other hand, if $\lambda_{i,i}$ is larger but starts to turn small, \textit{i.e.} $\mu \leq \lambda_{i,i} \leq \frac{\pi}{2}$, the subtractive margin $\mu$ will temporarily pull the positive video and text closer to regularize the gradient of the positive samples, thus restraining them from reaching perfect similarity. This intuition is formalized by the following theorem:


\begin{theorem}
Let $\lambda_{i,j}$ denote the angle between the representation of two samples $i, j$, $\mathcal{L}^{v,t}_{\textup{angular}, i}$ and $\mathcal{L}^{v,t}_{\textup{contrastive}, i}$ denote the training objectives with and without the angular margin, respectively. Then, if $\lambda_{i,i} \leq \frac{\pi}{2}$, the following inequality holds:
\begin{equation}
\small
\left|\frac{\partial \mathcal{L}_{\textup{angular},i}^{v,t}}{\partial \lambda_{i,i}}\right| \leq \left|\frac{\partial \mathcal{L}_{\textup{contrastive}, i}^{v,t}}{\partial \lambda_{i,i}}\right| 
\end{equation}
\end{theorem}
We provide the proof in Appendix \ref{app:proof_theorem_1}. In order to speed up the training in the beginning and constrain the update of positive pairs in the latter stage, we adopt an adaptive strategy that gradually increases the margin $\mu$ towards a limit:
\begin{equation}
\small
\mu^{(k)} = \frac{a_{0}}{a_{1} + e^{-a_{2} \cdot k}},
\end{equation}
where $k$ denotes the training step, and $a_{0}$, $a_{1}$, and $a_{2}$ denote hyperparameters.

\noindent\textbf{MLP-parameterized weighting function.} To control the effect of the data imbalance issue, we  construct a weighting function as a $\theta$-parameterized MLP to map each training loss to a sample weight:
\begin{equation}
\small
w(\theta, \mathcal{L}_{i}^{\text{train}}(\Theta)) = \text{MLP}_{\theta}(\mathcal{L}_{i}^{\text{train}}(\Theta)), \;\; i \in \{1, 2, …, B\}, 
\end{equation}
where $\mathcal{L}_{i}^{\text{train}}(\Theta)$ denotes the training loss of sample $i$ in a batch of size $B$, calculated using the $\Theta$-parameterized video-language model.

Then, the final training objective is the sum of the training losses of data samples weighted by $w$:
\vspace{-5pt}
\begin{equation}
\small
\mathcal{L}(\Theta, \theta) = \frac{1}{B}\sum\limits_{i=1}^{B} w(\theta, \mathcal{L}_{i}^{\text{train}}(\Theta)) \cdot \mathcal{L}_{i}^{\text{train}}(\Theta).
\vspace{-5pt}
\end{equation}
\noindent\textbf{Meta-optimized learning framework.} At present, we have a $\theta$-parameterized MLP network and a $\Theta$-parameterized video-language model to train. Since jointly training these two models might be unstable \citep{shu2019meta}, we instead develop a meta-learning approach. In our procedure, we manually extract a small amount of unbiased meta-data set $\{\textbf{v}_{i}^{\text{meta}}, \textbf{t}_{i}^{\text{meta}}, y_{i}^{\text{meta}}\}_{i=1}^{M}$, i.e. with semantically aligned textual description and balanced label distribution, representing the meta-knowledge of the groundtruth sample-label distribution, where $M$ is the number of the meta-samples and $M \ll N$. 

We first estimate the update of the video-language model parameters $\Theta$:
\begin{equation}
\small
\hat{\Theta}^{(k)} = \Theta^{(k)} - \frac{\alpha}{B} \sum\limits_{j=1}^{B} w_{i}\left(\theta^{(k)}, \mathcal{L}_{j}^{\text{train}}\left(\Theta^{(k)}\right)\right) \cdot \nabla_{\Theta} \mathcal{L}_{j}^{\text{train}}\left(\Theta^{(k)}\right),
\label{eq:estimate_video_language_model}
\vspace{-5pt}
\end{equation}
where $\alpha$ is the learning rate for the video-language model. Subsequently, we update the MLP network using the estimated $\hat{\Theta}$:
\begin{equation}
\small
\theta^{(k+1)} = \theta^{(k)} - \frac{\eta}{M} \sum\limits_{i=1}^{M} \nabla_{\theta} \mathcal{L}^{\text{meta}}_{i} \left(\hat{\Theta}^{(k)}\right),
\label{eq:update_mlp_network}
\vspace{-5pt}
\end{equation}
where $\beta$ is the learning rate for the MLP network. Lastly, we obtain the new video-language model’s parameters as:
\begin{equation}
\vspace{-5pt}
\begin{split}
\Theta^{(k+1)} = \Theta^{(k)} - \frac{\alpha}{B}\sum\limits_{j=1}^{B} w\left(\theta^{(k+1)}, \mathcal{L}_{j}^{\text{train}}\left(\Theta^{(k)}\right)\right) \cdot \nabla_{\Theta}\mathcal{L}^{\text{train}}_{j}\left(\Theta\right)\Big|_{\Theta^{(k)}}.
\end{split}
\label{eq:update_video_language_model}
\vspace{-5pt}
\end{equation}
We summarize our algorithm in Algorithm \ref{algo:meta_optimized_framework}. From an empirical perspective, we compare the effectiveness of joint learning and our meta-optimized framework in Section \ref{subsect:ablation_study}. From a theoretical perspective, we conduct further derivation for Eq. (\ref{eq:update_mlp_network}), resulting in the following formulation (more details in Appendix \ref{app:derivation_detail_update_rule_mlp}):
\begin{gather}
\theta^{(k+1)} = \theta^{(k)} + \frac{\alpha\beta}{BM} \sum_{i=1}^{M}\sum_{j=1}^{B} G_{ij} \frac{\partial w\left(\theta^{(k)}, \mathcal{L}_{j}^{\text{train}}\left(\Theta^{(k)}\right)\right)}{\partial\theta}\Big|_{\theta^{(k)}}, \\
G_{ij} = \nabla_{\hat{\Theta}} \mathcal{L}_{i}^{\text{meta}} \left(\hat{\Theta}\right)\Big|_{\hat{\Theta}^{(k)}} \cdot \nabla_{\Theta} \mathcal{L}_{j}^{\text{train}} \left(\Theta\right)\Big|_{\Theta^{(k)}},
\end{gather}
where $G_{ij}$ denotes the coefficient of the gradient. The coefficient will increase for training samples whose gradient is in the same direction with the gradient of the meta-data (meta gradient), as this is more likely to be a clean learning direction. In contrast, the coefficient will decrease for those whose gradient is opposite from the meta gradient, since the learning direction is likely to be noisy.
\vspace{-15pt}
\subsection{Large Vision-Language Model for Augmentation}
\vspace{-0.5mm}
To further enhance our video-language representation learning, we devise a strategy to utilize large vision-language model to augment additional video-text data. Our strategy consists of two stages, \textit{i.e.} extracting key frames and generating textual descriptions, which we delineate as follows:

\noindent\textbf{Key frame extraction.} Inspired by \citep{jin2023video}, we adopt a density peak-based clustering approach to extract key video frames. Starting with a sequence of visual frame features $\textbf{V}_{i}^{f} = \{\mathbf{h}^{f}_{i,j}\}_{j=1}^{N_{\textbf{v}_{i}}}$, we calculate the locality density $d_{i,j}$ of each feature $\mathbf{h}^{f}_{i,j}$ based on its $K$-nearest neighbors:
\begin{equation}
\vspace{-10pt}
\small
d_{i,j} = \exp\left(-\frac{1}{K} \sum\limits_{\textbf{h}^{f}_{i,l} \in \text{KNN}(\mathbf{h}^{f}_{i,j})} \left|\left|\textbf{h}^{f}_{i,l} - \textbf{h}^{f}_{i,j}\right|\right|^{2}\right),
\end{equation}
where $\text{KNN}(\mathbf{h}^{f}_{i,j})$ denotes the $K$-nearest neighbors of $\mathbf{h}^{f}_{i,j}$.

Subsequently, we estimate the distance index $\gamma_{i,j}$ of each frame $\mathbf{h}_{i,j}^{f}$:
\begin{align}
\small
\gamma_{i,j}=
\begin{cases}
\min\limits_{l: d_{i,j} > d_{i,l}} ||\mathbf{h}^{f}_{i,l} - \mathbf{h}^{f}_{i,j}||^{2},& \text{if } \exists l \; \text{s.t.} \; d_{i,l} > d_{i,j} \\
\max\limits_{l} ||\mathbf{h}^{f}_{i,l} - \mathbf{h}^{f}_{i,j}||,& \text{otherwise}
\end{cases}.
\end{align}

Our intuition is that $d_{i,j}$ denotes the local density of a video frame, and $\gamma_{i,j}$ denotes the distance from other frames of high density. We proceed to extract the top $Q$ video frames of the highest $d_{i,j} \times \gamma_{i,j}$ values as key video frames.

\noindent\textbf{Textual description generation.} 
Since large vision-language models have established impressive results on natural instruction tuning and visual reasoning capabilities, we leverage a LVLM to augment textual descriptions for video data \citep{dai2305instructblip, li2023blip, liu2023visual}. Particularly, given the top $Q$ video frames, we concatenate them into a single image $I$ of the $W \times H$ grid. We then forward the image $I$ along with the textual prompt ``\textit{Write a short caption sentence for the video in order from left to right, top to bottom}’’ to a large vision-language model to generate the sentence-level textual description $\textbf{t}$, which we combine with the video $\textbf{v}$ to form a video-text pair $(\textbf{v},\textbf{t})$. We illustrate the generated textual descriptions of our strategy in Appendix \ref{app:more_qualitative_results}.

%% file: files/04_experiments.tex
\vspace{-2mm}
\section{Experiments}
\vspace{-2mm}
\subsection{Experimental Settings}
\noindent\textbf{Downstream tasks.} Following previous works \citep{jin2023video, sun2019videobert, zhao2023learning, man2023tevl}, we evaluate our framework on two popular video question answering and text$\leftrightarrow$video retrieval tasks. We provide the details of the datasets in Appendix \ref{app:datasets}:
\begin{itemize}
    \item Video question answering (VideoQA): We experiment with two videoQA settings, \textit{i.e.} open-ended and multi-choice videoQA. Open-ended videoQA classifies a pair of video and question into one of the pre-defined set of answer labels. Multi-choice videoQA chooses the correct answer from five choices given the video and the language question. We assess our video-language representations by the VideoQA task using the following five datasets,  MSRVTT \citep{xu2016msr}, MSVD \citep{chen2011collecting}, TGIF-QA-R \citep{peng2021progressive}, NExT-QA \citep{xiao2021next}, and Causal-VidQA \citep{li2022representation}. 
    \item Text$\leftrightarrow$video retrieval: The retrieval task is to extract the corresponding video given the textual query, or extract the textual description given the video. We evaluate the retrieval ability of our video-language representations using three datasets: MSRVTT \citep{xu2016msr}, DiDeMo \citep{anne2017localizing}, ActivityNet \citep{krishna2017dense}.
\end{itemize}
\vspace{-5pt}
\noindent\textbf{Evaluation metrics.} For the videoQA task, we use the answer accuracy as the evaluation metric. For the retrieval task, we employ Recall at rank $K$ ($R@K$), with $K \in \{1, 5, 10\}$ to evaluate the performance.

\noindent\textbf{Video-language backbones.} To extensively validate the effectiveness of our framework, we conduct experiments on various models of bidirectional and dual architectures. Particularly, for the dual architecture, we experiment with CLIP-ViP \citep{xue2022clip} and video graph transformer (VGT) \citep{xiao2023contrastive}, while for the bidirectional architecture, we experiment with the well-known VIOLET architecture \citep{fu2023empirical}.

\noindent\textbf{Baseline methods.} We compare our method with a comprehensive list of video-language understanding models, along with the approaches that use LVLM to augment video-text data: (i) \textbf{CLIP4Clip} \citep{luo2022clip4clip}, a model that transfers image-text CLIP model \citep{radford2021learning} to text$\leftrightarrow$video retrieval tasks; (ii) \textbf{MERLOT} \citep{zellers2021merlot}, a method that trains on both spatial and temporal objectives to learn video-language representations; (iii) \textbf{LAVENDER} \citep{li2023lavender}, a model that learns video-language representations using a unified framework of masked language modeling; (iv) \textbf{Singularity} \citep{lei2022revealing}, curates a pre-training dataset and uses an early fusion strategy to improve single-frame video-language representation learning; (v) \textbf{OmniVL} \citep{wang2022omnivl}: a model that is trained on both image-language and video-language data to enhance video-language representation learning; (vi) \textbf{VindLU} \citep{cheng2023vindlu}, follows a recipe to select pre-training objectives, pre-training data, and model scale for effective video-language representation learning; (vii) \textbf{All-in-one} \citep{wang2023all}, a model that embeds raw video and textual signals into hidden representations without using pre-trained unimodal encoders; (viii) ( \textbf{DRL} \citep{wang2022disentangled}, a disentangled representation method that decouples sequential from hierarchical representations to specifically improve both of them for cross-modal retrieval; (ix) \textbf{CLIP2Video} \citep{bogolin2022cross}, consists of two normalization methods to improve the effectiveness and robustness of cross-modal retrieval models; (x) \textbf{LaViLa} \citep{zhao2023learning}, an approach that fine-tunes LLM to be conditioned on visual frames to generate additional textual descriptions, which are used to improve video-language representation learning; (xi) \textbf{Vid2Seq} \citep{yang2023vid2seq}, a pre-trained dense video captioning model which is used to generate additional textual descriptions for video data, which are employed to enhance video-language representation learning.

\noindent\textbf{Implementation details.} We utilize LLaVA model as the LVLM \citep{liu2023visual} to augment video inputs of the HowTo100M dataset \citep{miech2019howto100m}, which is a large-scale dataset for video-language representation learning but has been known for its weakly aligned video-text pairings \citep{han2022temporal}. To construct the input for the LVLM, we use $K = 6$ and $Q = 12$, then concatenate key frames into an image as a grid of $3\times4$ frames. Based upon validation, we adopt $\tau = 1$, $a_{0} = 0.2, a_{1} = 10$, and $a_{2} = -0.1$ for angular margin contrastive learning, and $\eta = 1$ for our optimization objective. For videoQA, to fairly compare with previous methods, we fine-tune our VIOLET-based model using AdamW with an initial learning rate of $\alpha = \beta = 2e-5$, betas of (0.9, 0.98), weight decay of 1e-3, and batch size $B$ of 4 for 10 epochs. Similarly, for the VGT-based model, we use Adam optimizer with an initial learning rate of 1e-5 of a cosine annealing schedule, and batch size $B$ of 4 for 30 epochs. For the text$\leftrightarrow$video retrieval task, we fine-tune the model with an initial learning rate of 5e-6 and a fixed weight decay of 5e-2, using a batch size $B$ of 4 for 5, 20, and 20 epochs on MSRVTT, DiDeMo, and ActivityNet datasets, respectively. More details can be found in Appendix \ref{app:implementation_details}.

\vspace{-10pt}
\subsection{Main Results}
\begin{table}[t]
\centering
\caption{VideoQA results on MSRVTT, MSVD, and TGIF-QA-R. Open-ended videoQA is evaluated on MSRVTT, MSVD, and TGIF-Frame datasets. Multi-choice videoQA is evaluated upon TGIF-Action and TGIF-Transition.}
\label{tab:videoqa_msrvtt_msvd_tgif_qa}
\resizebox{0.7\linewidth}{!}{
\begin{tabular}{l|c|c|ccc}
\hline
\multicolumn{1}{c|}{\multirow{2}{*}{\textbf{Method}}} & \multirow{2}{*}{\textbf{MSRVTT-QA}} & \multirow{2}{*}{\textbf{MSVD-QA}} & \multicolumn{3}{c}{\textbf{TGIF-QA-R}}                 \\
                               &                                     &                                   & \textbf{Action} & \textbf{Frame} & \textbf{Transition} \\ \hline
MERLOT                                              & 43.1                                &   51.9	                              &      61.4	            &         69.3	             &              84.0  \\
LAVENDER                                    & 44.2                                &             52.4                      &   57.1              &          66.9           &     84.0           \\
Singularity                                          & 43.5                                &         49.6                          &   53.1	              &       65.1              &         	81.5       \\
OmniVL                                            & 44.1                                &          51.0                         &     62.0	            &        69.5             &       85.5         \\
VindLU                                              & 44.6                                &         51.0                          &    59.5             &      65.8               &     87.2           \\
VGT                                                & 40.0                                &          36.4                         & 61.0            &       61.7              & 73.2           \\
VIOLET                                        &         44.5                            &         52.5                          &          62.6       &       70.0              &          86.3      \\
All-in-one                                          &         44.3                            &       47.9                            &       34.9          &     65.0                &       53.8         \\
LaViLa-VIOLET                                               &             44.9
                     &             53.7                      &     64.3		            &   70.7                  &          87.0      \\
Vid2Seq-VIOLET                                            &           44.8                             &  53.1                                 &     64.2		            &       70.1              &         86.6       \\ \hline
MAMA-VGT                               &    41.6	                                  &       37.1                            &      61.7		           &              62.7       &      74.0          \\
MAMA-VIOLET                            & \textbf{46.4}                                & \textbf{55.8}                             & \textbf{66.5}            & \textbf{71.7}                & \textbf{89.5}           \\ \hline
\end{tabular}}
\end{table}

\begin{table}[t]
\centering
\caption{VideoQA results on NExT-QA and Causal-VidQA. All of the datasets target multi-choice videoQA.}
\label{tab:videoqa_nextqa_causalvid_qa}
\resizebox{0.9\linewidth}{!}{
\begin{tabular}{l|cccc|cccc}
\hline
\multirow{2}{*}{\textbf{Method}} & \multicolumn{4}{c|}{\textbf{NExT-QA}}                                                                                                  & \multicolumn{4}{c}{\textbf{Causal-VidQA}}                                                                                                                                   \\ 
                              & \textbf{Descriptive} & \textbf{Temporal} & \textbf{Causal} & \textbf{All} & \textbf{Descriptive} & \textbf{Explanatory} & \textbf{Predictive} & \textbf{Counterfactual} \\ \hline
MERLOT                                             &        66.6			              &                        62.5	               &                              58.2	       &           59.4                       &            67.2                              &                   65.7	                       &                         57.2	                &          57.0                                   \\
LAVENDER                                           &     63.4                 &  56.5                                      &     54.6                                &          56.7                        &         62.0                                 &          61.6                                &     46.3                                    &     50.4                                        \\
Singularity                                          &      60.0						                &        61.1                               &    50.9                                 &          54.8                        &   54.3                                       &                  49.6                        &            41.3                             &                  46.9                           \\
OmniVL                                                &          67.1	            &       63.8                             &         55.9                             &             59.9                     &             67.0                             &                  66.0                        &                       56.5                  &        57.1                                     \\
VindLU                                               &          68.4       &      59.7                                 &                55.7                      &   59.8                             &        63.6			                                  &           54.8                               &       57.7                                  &                 54.4                            \\
VGT                                              &     69.6                &     65.4            & 56.2            &      61.8      & 74.4                 & 75.6                 & 60.7                & 65.6                    \\
VIOLET                                         &          67.7           &    58.0                                   &       50.7                              &              58.5                    &      67.6			                                    &        66.6                                  &         57.1                                &         57.6                                    \\
All-in-one                                        &   	64.8							                   &       63.9                               &           50.9                          &          57.9                      &             60.4                          &                        51.3                   &                       51.2                     &                         50.7                     \\
LaViLa-VGT                                              &          		70.9     				            &                69.2                    &                            59.3       &                        64.8          &                            74.7	              &       76.1                                   &        61.6                                &               65.7                              \\
Vid2Seq-VGT                                             &   70.3					                   &                        67.0	               &          58.5                           &             64.2                     &                         73.7                 &          76.0                                &                    61.3	                     &                   65.6                          \\ \hline
MAMA-VGT                               &  \textbf{72.7}				                 &    \textbf{70.6}	              &        \textbf{62.2}         &     \textbf{66.3}       &                \textbf{75.3}     &     \textbf{77.1}	        &    \textbf{62.2}	                  &         \textbf{68.2}               \\
MAMA-VIOLET                            & 71.2                 & 68.6              & 61.1            & 65.8         & 72.7                 & 68.0                 & 60.5                & 59.5   \\ \hline
\end{tabular}}
\vspace{5pt}
\end{table}

\begin{table}[t]
\centering
\caption{Text$\rightarrow$video retrieval results on MSRVTT, DiDeMo, and ActivityNet.}
\label{tab:text_video_retrieval_msrvtt_didemo_activitynet}
\resizebox{0.8\linewidth}{!}{
\begin{tabular}{l|ccc|ccc|ccc}
\hline
\multirow{2}{*}{\textbf{Method}} & \multicolumn{3}{c|}{\textbf{MSRVTT}}         & \multicolumn{3}{c|}{\textbf{DiDeMo}}         & \multicolumn{3}{c}{\textbf{ActivityNet}}    \\
                                 & \textbf{R@1} & \textbf{R@5} & \textbf{R@10} & \textbf{R@1} & \textbf{R@5} & \textbf{R@10} & \textbf{R@1} & \textbf{R@5} & \textbf{R@10} \\ \hline
VIOLET                          & 37.2         & 64.8         & 75.8          & 47.9         & 76.5         & 84.1          &   18.1           &      43.1        &    56.7           \\
VindLU                          & 48.8         & 72.4         & 82.2          & 59.8         & 86.6         & 91.5          & 55.9         & 82.3         & 90.9          \\
CLIP4Clip                        & 44.5         & 71.4         & 81.6          & 42.8         & 68.5         & 79.2          & 40.5         & 72.4         & 83.4          \\
CLIP2Video                     & 47.2         & 73.0         & 83.0          &     -         &        -      &      -         & -            & -            & -             \\
DRL                             & 47.4         & 74.6         & 83.8          & 47.9         & 73.8         & 82.7          & 44.2         & 74.5         & 86.1          \\
CLIP-ViP                         & 55.7         & 77.7         & 86.6          & 55.7         & 78.1         & 86.1          & 59.1         & 83.9         & 91.3          \\ 
LaViLa - CLIP-ViP                      &   56.0       &     79.8     &      87.2     &      56.6    &    79.8      &   87.1        &      58.7    &    82.8      &     90.5      \\ 
Vid2Seq - CLIP-ViP                      &      55.3	     &    77.9      &    86.1       &   57.6       &   79.9       &    88.4       &       55.1    &     79.0     &     87.4       \\ \hline
MAMA - CLIP-ViP      & \textbf{60.0}         & \textbf{82.2}         & \textbf{89.2}          &    \textbf{62.7}          &       \textbf{89.9}       &     \textbf{96.0}          & \textbf{60.0}         & \textbf{85.0}         & \textbf{91.7}       \\ \hline
\end{tabular}}
\end{table}

\begin{table}[h!]
\centering
\caption{Video$\rightarrow$text retrieval results on MSRVTT, DiDeMo, and ActivityNet.}
\label{tab:video_text_retrieval_msrvtt_didemo_activitynet}
\vspace{-10pt}
\resizebox{0.7\linewidth}{!}{
\begin{tabular}{l|ccc|ccc|ccc}
\hline
\multirow{2}{*}{\textbf{Method}} & \multicolumn{3}{c|}{\textbf{MSRVTT}}         & \multicolumn{3}{c|}{\textbf{DiDeMo}}         & \multicolumn{3}{c}{\textbf{ActivityNet}}    \\
                                 & \textbf{R@1} & \textbf{R@5} & \textbf{R@10} & \textbf{R@1} & \textbf{R@5} & \textbf{R@10} & \textbf{R@1} & \textbf{R@5} & \textbf{R@10} \\ \hline
VIOLET                          &     36.6		         &        64.1      &       75.1         &     44.8    &      72.9        &        82.4       &      15.8        &      39.8        &      54.8         \\
VindLU                          &     46.0	         &   71.8           &        80.2       &         57.2		     &       83.1       &        88.9       &     47.1	        &    77.5          &      86.6         \\
CLIP4Clip                        & 43.1         & 70.5         & 81.2          &   17.5           &    37.5          &       49.4        & 42.6         & 73.4         & 85.6          \\
CLIP2Video                     &  43.5		            &      72.3        &         82.1       &       -      &        -      &     -          &         -     &       -       &       -        \\
DRL                          &       45.1		       &     72.9            &   83.5         &   45.4     &  72.6         &   82.1           &      42.2          &           74.0   &    86.2           \\
CLIP-ViP                         & 48.0         & 72.4         & 82.9          & 46.3         & 73.2         & 81.9          & 50.2         & 78.3         & 87.5          \\ 
LaViLa - CLIP-ViP                     &     49.0     &   74.5       &  83.4         &     47.1     &      74.2    &    83.0       &   50.3       &  78.7        &     88.4      \\ 
Vid2Seq - CLIP-ViP                        &  49.1			       &   74.1       & 82.5          &    47.2      &    74.4      &   83.1        &    50.9		     & 79.5         &      89.6    \\  \hline
Our method - CLIP-ViP      &   \textbf{50.1}         &   \textbf{76.8}         &   \textbf{84.8}          &   \textbf{48.1}         &   \textbf{78.1}         &   \textbf{85.5}          &   \textbf{52.3}         &   \textbf{80.8}         &   \textbf{90.1}        \\ \hline 
\end{tabular}}
\vspace{5pt}
\end{table}

We denote the results for videoQA in Table \ref{tab:videoqa_msrvtt_msvd_tgif_qa} and \ref{tab:videoqa_nextqa_causalvid_qa}, for text$\rightarrow$video retrieval in Table \ref{tab:text_video_retrieval_msrvtt_didemo_activitynet}, and for video$\rightarrow$text retrieval in Table \ref{tab:video_text_retrieval_msrvtt_didemo_activitynet}.

\noindent\textbf{VideoQA.} In terms of open-ended videoQA, we substantially outperform the LVLM approaches, \textit{e.g.} improve upon LaViLa with improvements of 1.5\% on MSRVTT, 2.1\% on MSVD, and 3.7\% on TGIF-Frame. We also surpass previous video-language understanding models, \textit{e.g.} VindLU by 1.8\% on MSRVTT, VIOLET by 3.3\% on MSVD, and OmniVL by 2.2\% on TGIF-Frame. Moreover, for multi-choice videoQA, we consistently surpass both LVLM and video-language understanding models, \textit{e.g.} achieving an overall gain of 1.5\% on NExT-QA and 1.2\% on Causal-VidQA over LaViLa, while the gains over VGT are 4.5\% and 1.6\%, respectively on these two datasets.

\noindent\textbf{Text$\leftrightarrow$video retrieval.} Our observation in videoQA applies for the text$\leftrightarrow$video retrieval task. In the text$\rightarrow$video direction, we enhance upon CLIP-ViP by 2.6 R@10 points on MSRVTT, 5.5 R@1 points on DiDeMo, and 1.1 R@5 points on ActivityNet. We also significantly outperform the LVLM approaches, \textit{e.g.} by 4.0 R@1 points and by 2.4 R@5 points on MSRVTT over LaViLa. Our superiority on video$\rightarrow$text retrieval is analogous to the text$\rightarrow$video case. 

These results substantiate that our framework is applicable to various video-language understanding tasks and model architectures. We hypothesize that we can better control the alignment between the positive video and language, and their effect upon the model training, thus leading to more reasonable video-language representations.

\vspace{-4mm}
\subsection{Ablation Study}
\label{subsect:ablation_study}
\vspace{-2mm}
We ablate our meta-optimized framework to investigate which factor results in the demonstrated effectiveness and explore intriguing properties. We conduct all experiments of our ablation study on the MSRVTT and MSVD for videoQA, and MSRVTT and DiDeMo for the text$\leftrightarrow$video retrieval task.

\begin{figure}[t]
\centering
\caption{(Left) Validation videoQA accuracy on MSVD with respect to $\mu$; (Middle) Validation videoQA accuracy on MSVD with respect to $a_0$; (Right) Relationship between loss values and weight values generated by our MLP-parameterized weighting function.}
\label{fig:analysis_mu_a0_weight}
\begin{subfigure}{0.30\linewidth}
    \centering
  \includegraphics[width=0.9\linewidth]{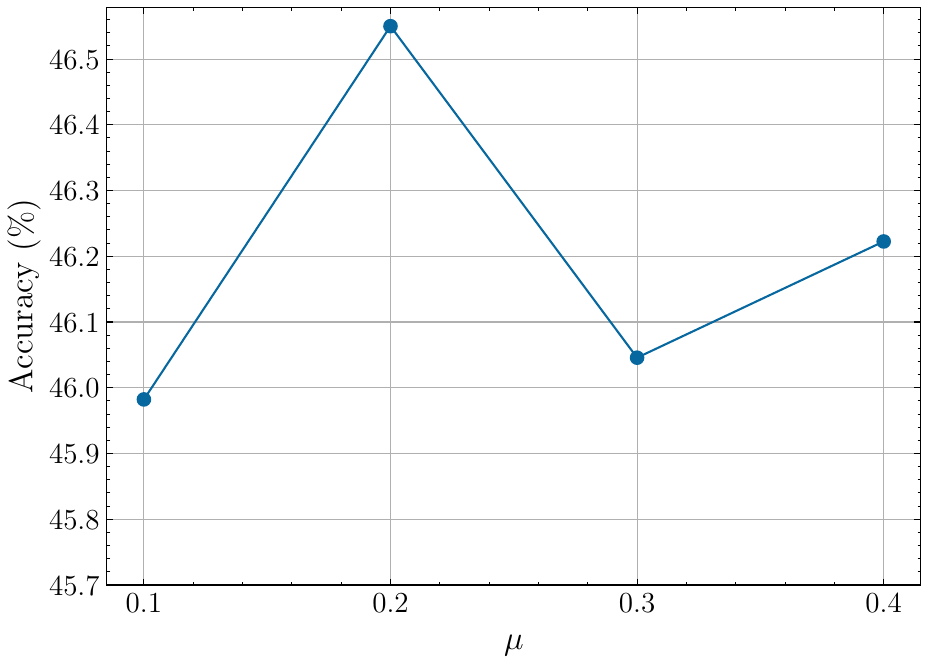} 
\end{subfigure}
\begin{subfigure}{0.30\linewidth}
    \centering
  \includegraphics[width=0.9\linewidth]{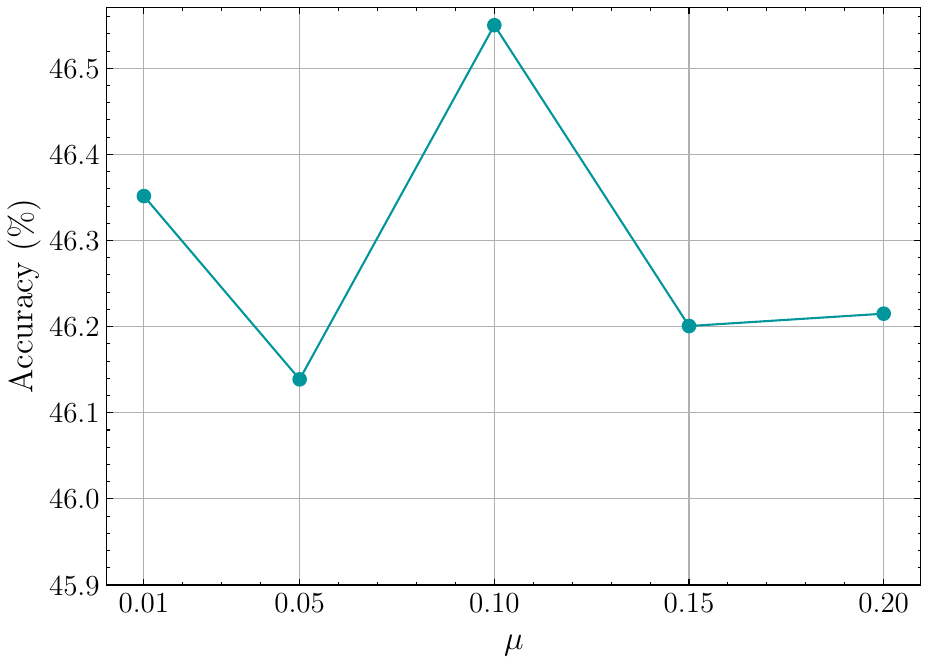} 
\end{subfigure} 
\begin{subfigure}{0.30\linewidth}
    \centering
  \includegraphics[width=0.9\linewidth]{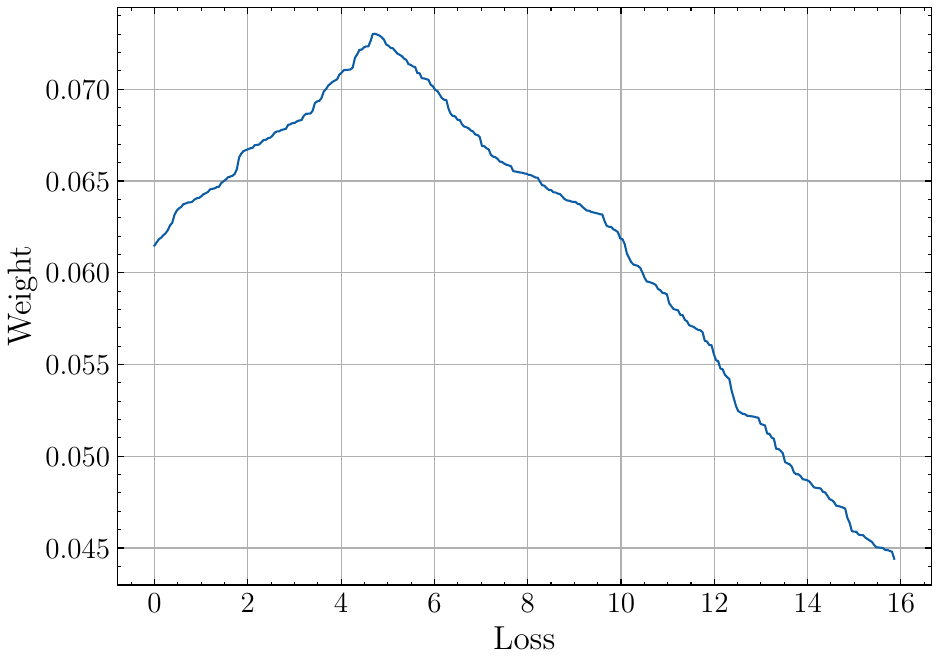} 
\end{subfigure} 
\end{figure}

\noindent\textbf{Varying the angular margin.} The angular margin, represented as $\mu$, is critical to control the relationship between semantically close video and language inputs. To better understand $\mu$, we experiment with manually varying the margin $\mu$ from 0.1 to 0.4 in Figure \ref{fig:analysis_mu_a0_weight} (left). We observe that the optimal performance is achieved when $\mu \sim 0.2$. Based upon this observation, we adopt $a_0 = 2$ and $a_{1} = 10$. Subsequently, we evaluate the impact of $a_2 \in \{0.01, 0.05, 0.1, 0.15, 0.2\}$ in Figure \ref{fig:analysis_mu_a0_weight} (right). We discover that $a_{2} = 0.1$ achieves the best performance and we adopt it for all experiments.

\noindent\textbf{Reweighting strategies.}  In addition to MLP, there exist various approaches to weigh training losses to control noisy pairings and data imbalance issues within the training data, \textit{\textit{e.g.}} focal loss \citep{lin2017focal}, self-paced learning (SPL) \citep{jiang2014self}, and L2RW \citep{ren2018learning}. We replace our MLP by these approaches and demonstrate the results in Table \ref{tab:exp_weighting_functions}. We find that our MLP-parameterized function significantly improves upon previous weighting functions that use manually designed formulation, \textit{i.e.} focal loss and SPL, and also the meta-learning approach without using MLP, \textit{i.e.} L2RW. The possible reason could be that MLP is a universal approximator, so that it can adaptively learn a reasonable weighting function based on data, thus outperforming methods that do not employ MLP.

\begin{table}[t]
\centering
\caption{Experiments on the LVLM choice for our prompting strategy.}    
\vspace{-5pt}
\label{tab:exp_lvlm_choice}
\resizebox{0.6\linewidth}{!}{
\begin{tabular}{l|cc|ccc|ccc}
\hline
\multirow{3}{*}{\textbf{LVLM choice}} & \multicolumn{2}{c|}{\textbf{VideoQA}}                              & \multicolumn{6}{c}{\textbf{Text$\rightarrow$video retrieval}}                           \\ \cline{2-9}
                                      & \multirow{2}{*}{\textbf{MSRVTT}} & \multirow{2}{*}{\textbf{MSVD}} & \multicolumn{3}{c|}{\textbf{MSRVTT}}         & \multicolumn{3}{c}{\textbf{DiDeMo}}         \\
                                      &                                  &                                & \textbf{R@1} & \textbf{R@5} & \textbf{R@10} & \textbf{R@1} & \textbf{R@5} & \textbf{R@10} \\ \hline
No augmentation                              &           45.6	                     &            54.7                    & 59.0		         &    80.3        &       87.6         &     58.0		        &  82.9            &    89.6          \\
BLIP-2                               &             45.7	                     &            54.9                    &     59.1		         &    80.4          &       87.7         &     58.1		        &  82.9            &    89.7           \\
InstructBLIP                     &              46.1	                    &             55.2                   &     59.2	         &      80.9	        &        88.1       &  58.5	             &        	83.5      &      89.8         \\
LLaVA-ChatGPT                                 &              46.1	               &           55.0             &      59.0		    &    80.3     &      87.7     &  58.7		    &  83.9     &  90.0   \\ \hline
LLaVA                                & \textbf{46.4}                             & \textbf{55.8}                           & \textbf{60.0}         & \textbf{82.2}         & \textbf{89.2}          & \textbf{62.7}         & \textbf{89.9}         & \textbf{96.0}         \\ \hline
\end{tabular}}
\end{table}

\begin{table}[t]
\centering
\vspace{-10pt}
\caption{Experiments on weighting functions.}
\label{tab:exp_weighting_functions}
\resizebox{0.6\linewidth}{!}{
\begin{tabular}{l|cc|ccc|ccc}
\hline
\multirow{3}{*}{\textbf{Weighting function}} & \multicolumn{2}{c|}{\textbf{VideoQA}}                              & \multicolumn{6}{c}{\textbf{Text$\rightarrow$video retrieval}}                                                                         \\  \cline{2-9}
                                             & \multirow{2}{*}{\textbf{MSRVTT}} & \multirow{2}{*}{\textbf{MSVD}} & \multicolumn{3}{c|}{\textbf{MSRVTT}}                                & \multicolumn{3}{c}{\textbf{DiDeMo}}                                \\ 
                                             &                                  &                                & \textbf{R@1}         & \textbf{R@5}         & \textbf{R@10}        & \textbf{R@1}         & \textbf{R@5}         & \textbf{R@10}        \\ \hline
Focal loss                                  &      45.0                            &         54.1                       &      58.3                &         80.0             &              86.9        &        59.7              &    85.6                  &        91.8              \\
SPL                                        &           45.3                       &    54.3                            &         58.7        &      80.8                &             88.1         &          61.4            &      87.5                &             93.8         \\
L2RW                                         &         45.9                         &        55.2                        &     59.5	            &        81.8              &      89.6                &             62.0         &       88.8               &      94.6                \\ \hline
MLP                                          &        \textbf{46.4}                                                        &          \textbf{55.8}                       &  \textbf{60.0}         & \textbf{82.2}         & \textbf{89.2}          &    \textbf{62.7}          &       \textbf{89.9}       &     \textbf{96.0} \\ \hline
\end{tabular}}
\vspace{5pt}
\end{table}

\noindent\textbf{Optimization strategies.} We explore different strategies to optimize the MLP and video-language model. Particularly, we experiment with the joint learning strategy, which simultaneously updates the parameters of the MLP and the video-language model to minimize the objective in Eq. \ref{eq:total_training_objective}. As shown in Table \ref{tab:exp_learning_strategy}, our meta-learning approach outperforms the joint learning one and achieves the best performance. This empirically validates the effectiveness of our meta-learning strategy which enables the training to follow gradient of the meta-data.

\begin{table}[t]
\centering
\caption{Experiments on the number of key frames $Q$ and the concatenated grid.}
\label{tab:exp_q_grid}
\resizebox{0.5\linewidth}{!}{
\begin{tabular}{c|c|cc|ccc|ccc}
\hline
\multirow{3}{*}{$Q$} & \multirow{3}{*}{\textbf{Grid}} & \multicolumn{2}{c|}{\textbf{VideoQA}}                                                                      & \multicolumn{6}{c}{\textbf{Text$\rightarrow$video retrieval}}                                                                                                                                                   \\ \cline{3-10}
                            &                                & \multirow{2}{*}{\textbf{MSRVTT}} & \multirow{2}{*}{\textbf{MSVD}} & \multicolumn{3}{c|}{\textbf{MSRVTT}}                                                                     & \multicolumn{3}{c}{\textbf{DiDeMo}}                                                                     \\
                            &                                &                                  &                                & \textbf{R@1} & \textbf{R@5} & \textbf{R@10} & \textbf{R@1} & \textbf{R@5} & \multicolumn{1}{c}{\textbf{R@10}} \\ \hline
1                           & $1\times1$                            &            45.0	                    &     55.1                           &    58.2		          &         79.3     &     86.2             &    60.2		   &     85.8       &   93.2          \\ \hline
\multirow{2}{*}{2}          & $1\times2$                            &             45.7	                     &        55.3                        &     	 58.8 	        &      80.0            &     87.0          &   60.5         &    85.8        &   93.3            \\
                            & $2\times1$                            &                      45.9	                                &       55.4                                             &          58.4                        &          79.2                     &               87.0                     &           60.9		                       &             86.1                     &                 94.2                  \\ \hline
\multirow{2}{*}{4}          & $1\times4$                            &           45.9	                       &     55.2                           &   59.1		 	         &   	    80.3      &       88.4          &   61.3		       &    86.3        &        94.6       \\
                            & $2\times2$                            &        45.6	                                              &                55.2                                    &       59.2                            &     80.4                             &          88.4                          &       61.3		                           &             86.7                     &           94.6                        \\ \hline
\multirow{2}{*}{8}          & $2\times4$                           &       45.9	                                               &       55.6                                             &             			59.8		             &     81.3                 &              88.7                    &        61.8		                          &                      89.0             &                     95.3            \\
                            & $4\times2$                            &                45.8	                                      &              55.5                                      &            59.3	                      &        80.7           &       88.5                       &     61.8		                             &           88.9                      &              95.3                       \\ \hline
\multirow{2}{*}{12}         & $3\times4$                            &      \textbf{46.4}                                                        &            \textbf{55.8}                       &    \textbf{60.0}         &   \textbf{82.2}         &   \textbf{89.2}          &      \textbf{62.7}          &         \textbf{89.9}       &       \textbf{96.0}                              \\
                            & $2\times6$                           &                   45.8                                   &           	55.6                                         &       59.3	                           &             81.7	                     &                 89.1                  &   62.7	                               &            	89.2                      &      95.6                             \\ \hline
\multirow{2}{*}{16}         & $2\times8$                            &                   45.8	                                   &             55.1                                       &                58.3		                  &              79.7                    &       87.6                            &              61.9		                    &               88.7                   &           94.8                        \\
                            & $4\times4$                            &                   45.7	                                   &              55.1                                      &         58.9		                         &             80.8                     &        88.7                           &         62.6	                        &              89.1                    &       95.9                             \\ \hline
\end{tabular}}
\end{table}

\noindent\textbf{Video-text data augmentation.} We assess the influence of different video-text data augmentation strategies. In particular, we ablate the LLaVA-generated samples and also explore different LVLM choices, \textit{i.e.} BLIP-2 \citep{li2023blip} and InstructBLIP \citep{dai2305instructblip}. In addition, we also use LLaVA model to generate image caption for each video frame, and ask ChatGPT to write a summary for the concatenated captions of all frames as the textual description. Table \ref{tab:exp_lvlm_choice} shows that neglecting LVLM-generated data results in slight performance degradation, which can be resolved by using other LVLMs, but still lack behind LLaVA model. We conjecture that LLaVA can better follow the language instruction to produce more accurate textual descriptions for videos \citep{liu2023visual}. Moreover, LLaVA-ChatGPT approach also leads to performance degradation, possibly because it is still challenging for ChatGPT to infer temporal relations among captions of separate video frames.

\begin{table}[t]
\centering
\caption{Experiments on the learning strategy.}
\label{tab:exp_learning_strategy}
\resizebox{0.6\linewidth}{!}{
\begin{tabular}{l|cc|ccc|ccc}
\hline
\multirow{3}{*}{\textbf{Training strategy}} & \multicolumn{2}{c|}{\textbf{VideoQA}}                              & \multicolumn{6}{c}{\textbf{Text$\rightarrow$video retrieval}}                           \\ \cline{2-9}
                                            & \multirow{2}{*}{\textbf{MSRVTT}} & \multirow{2}{*}{\textbf{MSVD}} & \multicolumn{3}{c|}{\textbf{MSRVTT}}         & \multicolumn{3}{c}{\textbf{DiDeMo}}         \\
                                            &                                  &                                & \textbf{R@1} & \textbf{R@5} & \textbf{R@10} & \textbf{R@1} & \textbf{R@5} & \textbf{R@10} \\ \hline
Joint learning                              &               45.5                   &      54.9                          &    59.3          &  81.1            &        88.4       &       59.2       &       84.9       &      90.2         \\
Meta-learning                               &        \textbf{46.4}                                                        &          \textbf{55.8}                       &  \textbf{60.0}         & \textbf{82.2}         & \textbf{89.2}          &    \textbf{62.7}          &       \textbf{89.9}       &     \textbf{96.0}      \\ \hline  
\end{tabular}}
\end{table}

\begin{figure*}[t]
    \centering
    \caption{a) Relative R@1 improvement in the text-video retrieval task on MSRVTT for each topic; b) Relative accuracy improvement in the videoQA task on MSRVTT with respect to the proportion of questions for which each label is the answer.}
    \label{fig:improvement_figure}
    \vspace{-10pt}
    \begin{subfigure}[h!]{0.45\linewidth}
        \centering
        \includegraphics[width=\linewidth]{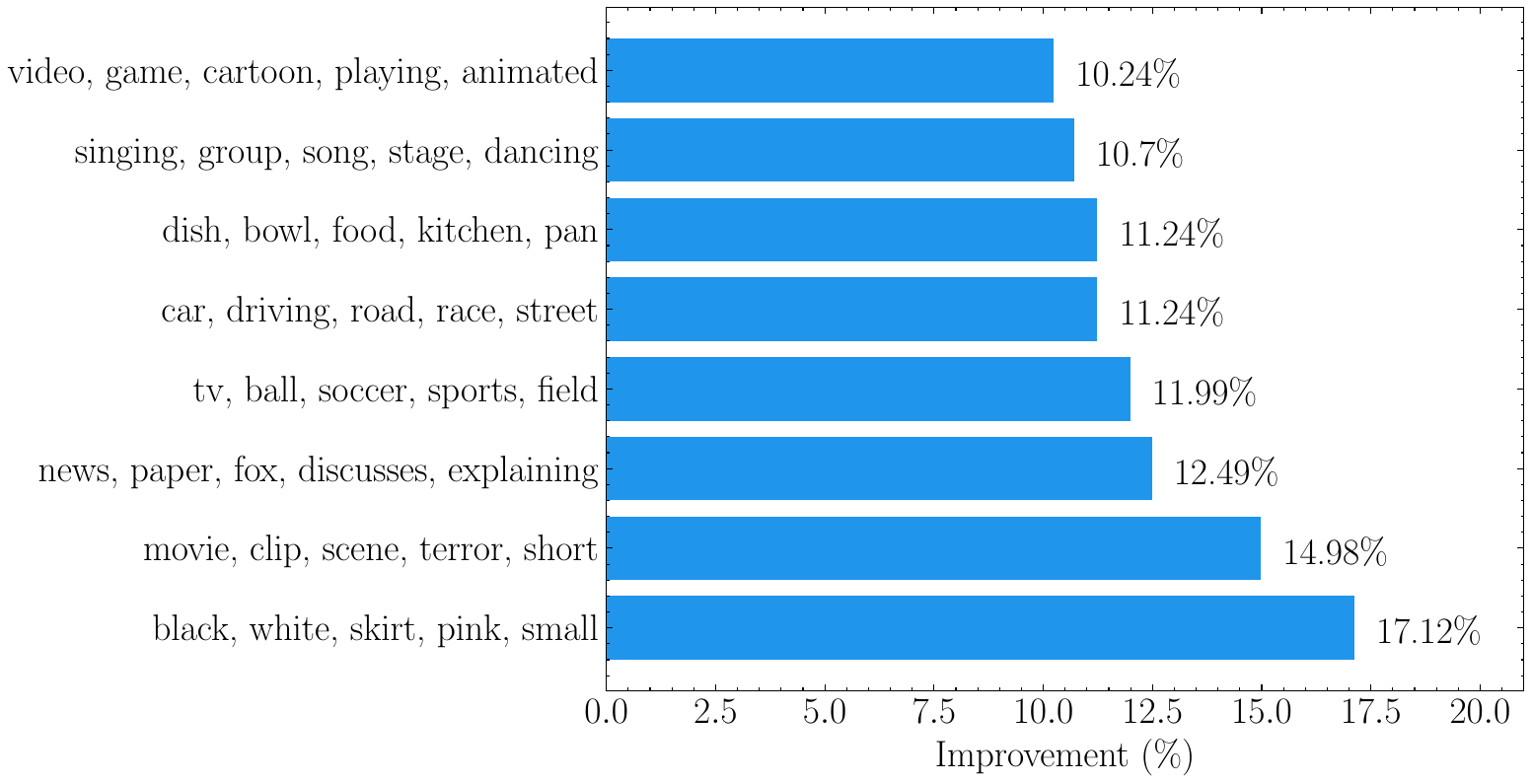}
    \end{subfigure}
    \hspace{3mm}
    \begin{subfigure}[h!]{0.35\linewidth}
        \centering
        \includegraphics[width=0.9\linewidth]{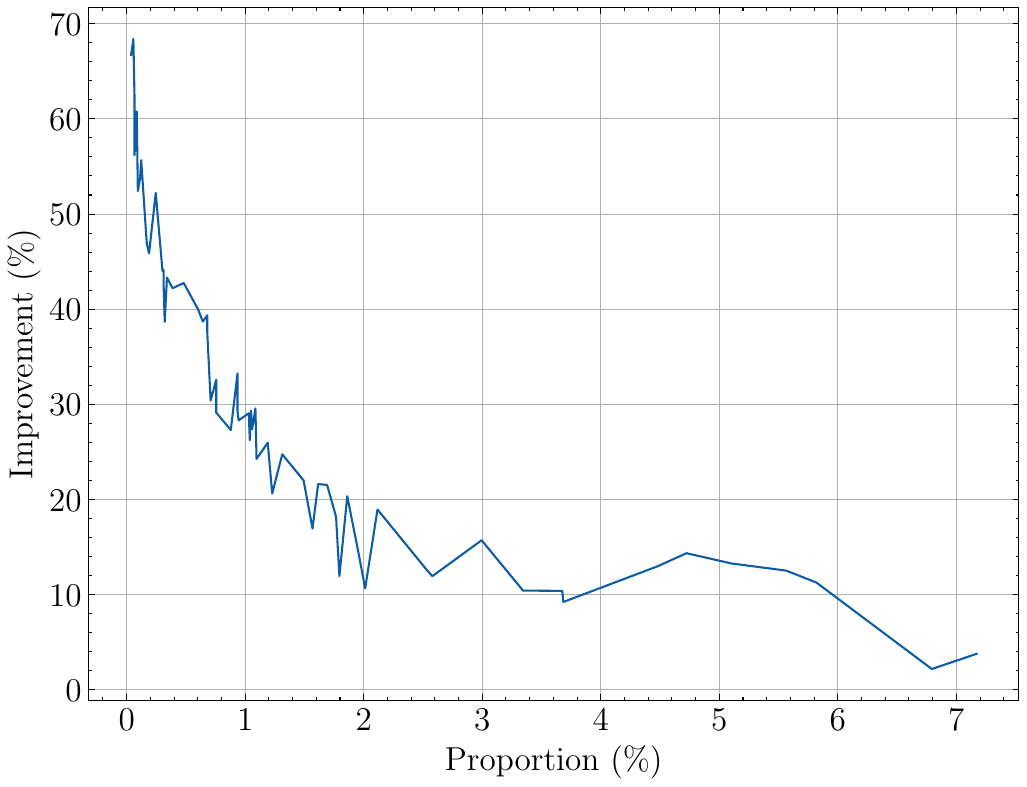}
    \end{subfigure}
\vspace{-10pt}
\end{figure*}

\begin{table}[h!]
\centering
\caption{Case study of similarity scores for video-text pairs.}
\vspace{-10pt}
\label{tab:similarity_score_video_text_pairs}
\resizebox{0.8\linewidth}{!}{
\begin{tabular}{c|p{0.7\linewidth}|c|c}
\hline
\textbf{Video} & \multicolumn{1}{c|}{\textbf{Caption}} & \textbf{CLIP-ViP score} & \textbf{Our score} \\ \hline
\multirow{4}{*}{\includegraphics[width=0.1\linewidth]{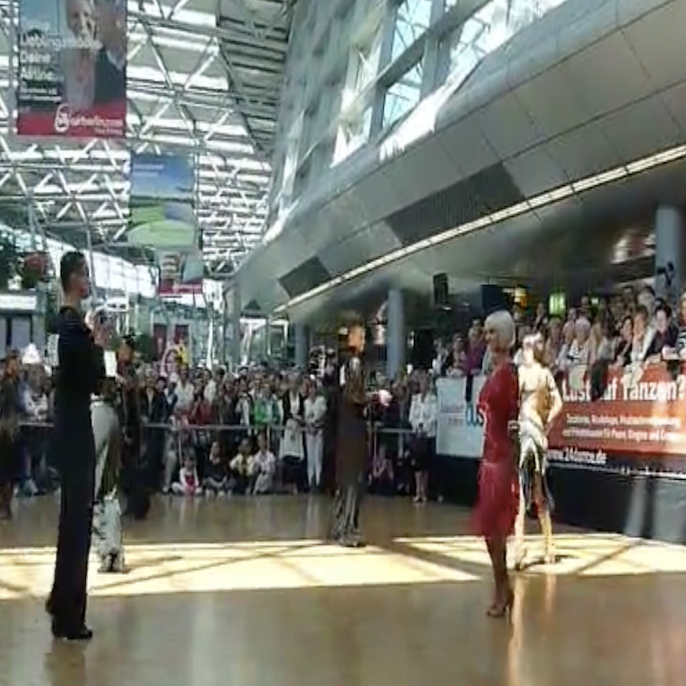} \includegraphics[width=0.1\linewidth]{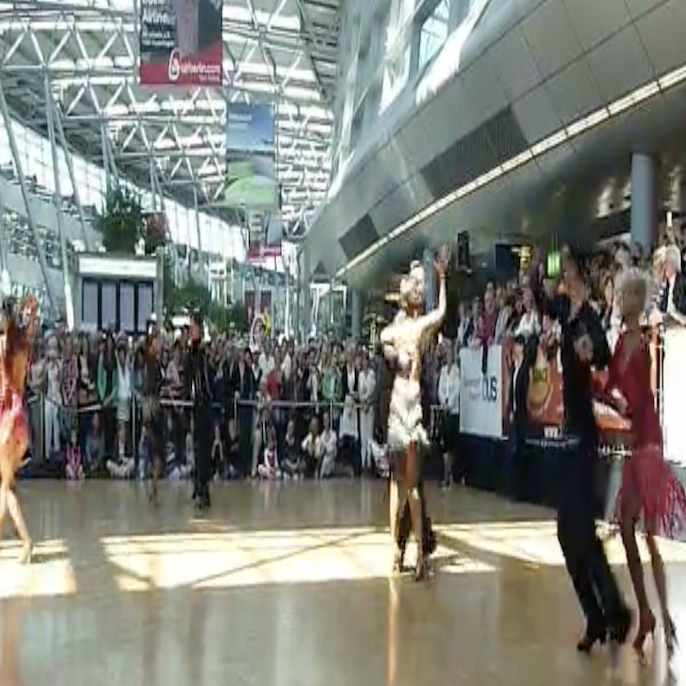}
\includegraphics[width=0.1\linewidth]{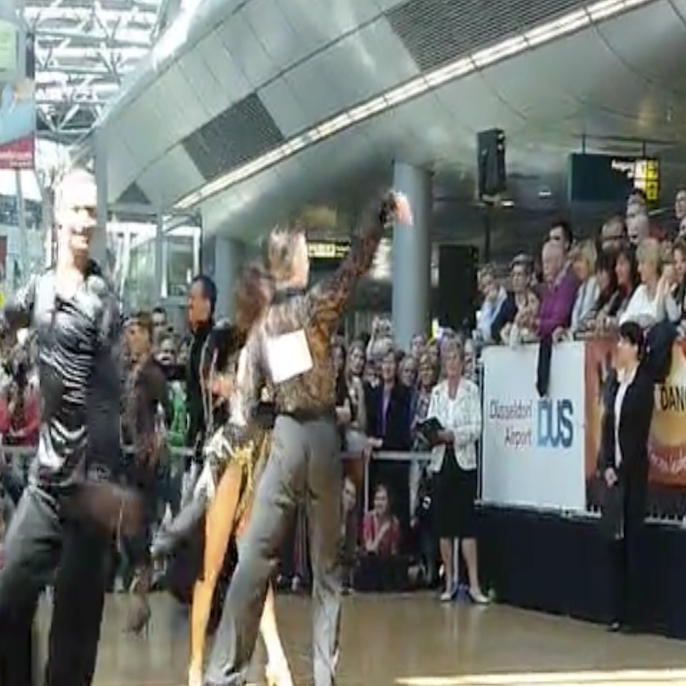} \includegraphics[width=0.1\linewidth]{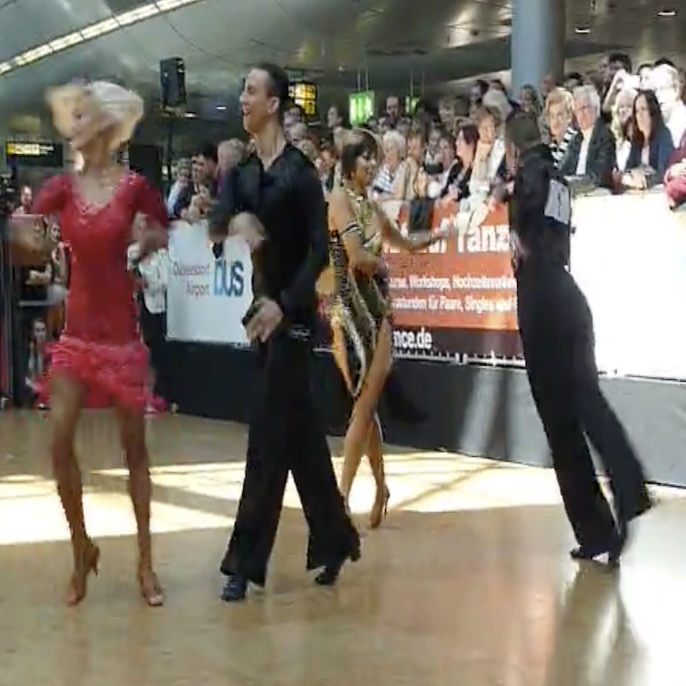}}                  &      a group of people dancing on a stage in front of a crowd.                                &    0.1727                                         &            0.1698                            \\ \cline{2-4}
                                   &      the man in black reaches his dance partner. man walks to the woman in red the camera zooms out on the dancers. we see a dancer in the back left spinning                                &   0.2175                                          &  0.3342                                      \\ \hline
\multirow{5}{*}{\includegraphics[width=0.1\linewidth]{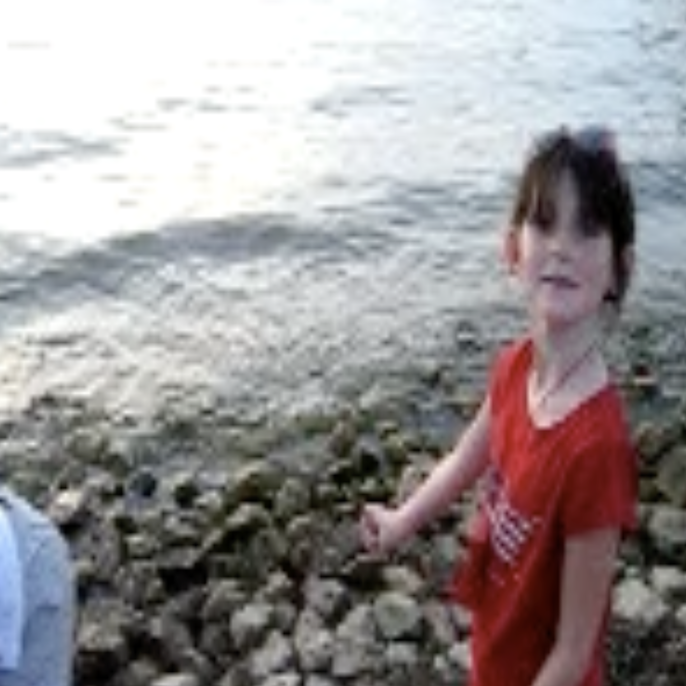} \includegraphics[width=0.1\linewidth]{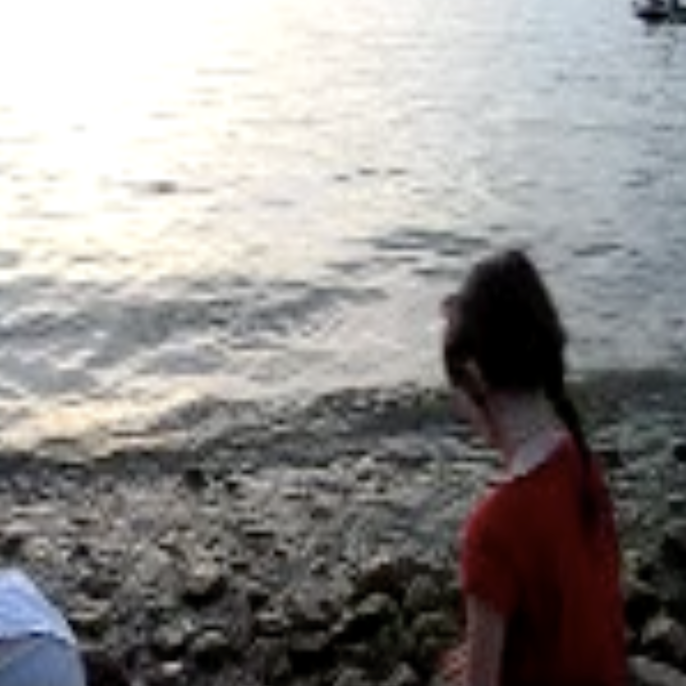}
\includegraphics[width=0.1\linewidth]{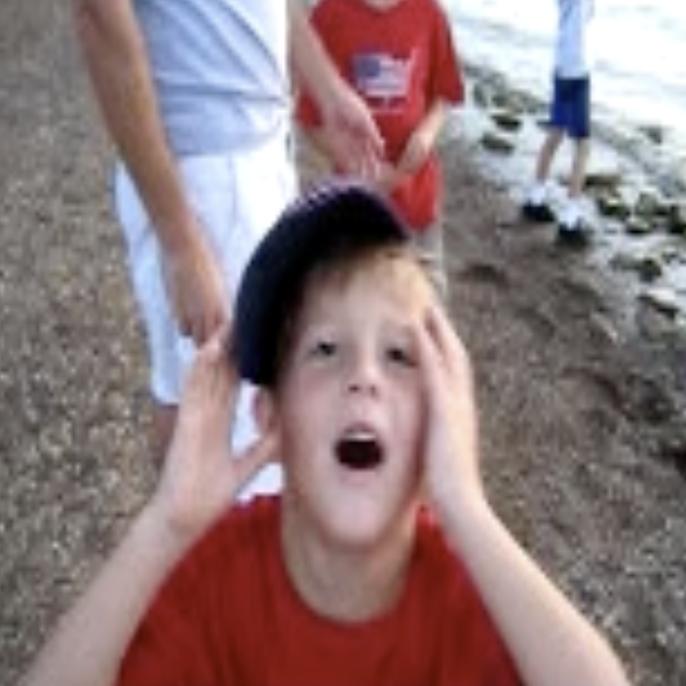} \includegraphics[width=0.1\linewidth]{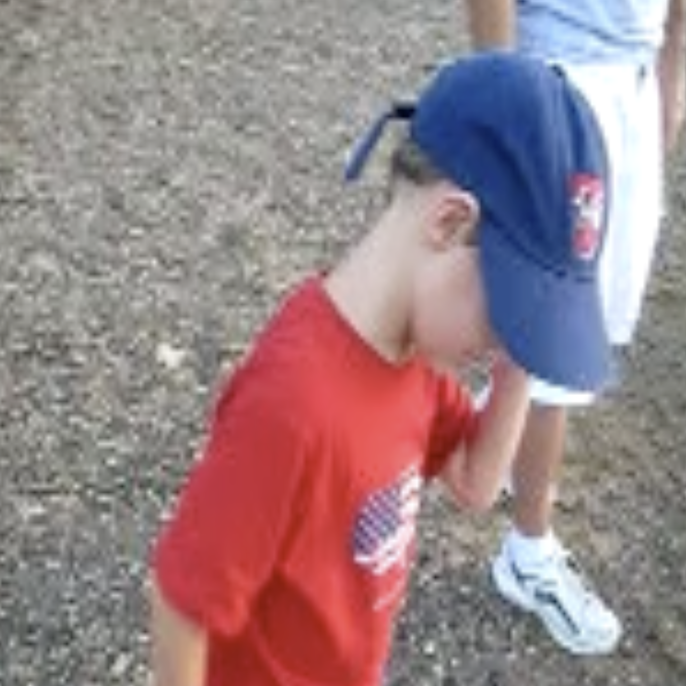}}                  &              a boy in a red shirt and a blue hat is playing.                        &       0.2594                                      &              0.2441                          \\ \cline{2-4}
                                   &     a girl throwing a rock. girl in the red shirt looks back over his shoulder the boy in a blue hat is seen for the first time. the boy in the baseball cap first looks at the camera boy wearing a blue ball cap                                 &   0.2869                                          &   0.3164  \\    \hline
\end{tabular}}
\vspace{5pt}
\end{table}

\noindent\textbf{Number of extracted key frames.} We explore the impact of varying number of extracted key frames and varying the approach to concatenate in Table \ref{tab:exp_q_grid}. While the performance is quite invariant to the concatenation strategy, we observe an increasing trend of the performance when the number of the extracted frames $Q$ increases, since there exists more information for LVLM to put into the descriptions. However, after surpassing $Q = 12$ frames, the performance dramatically decreases. The reason could be that when $Q$ is excessively large, visual frames become excessively small for LVLM to precisely detect details. Furthermore, balanced grids, \textit{i.e.} $3 \times 4$ and $4 \times 4$, tend to outperform skewed grids, \textit{i.e.} $2 \times 6$ and $2 \times 8$, respectively. The cause might be that video details are more difficult for an LVLM to capture when the video frames are presented on a long rectangle than when they are presented on a balanced one.


\vspace{-15pt}
\subsection{Analysis}
\noindent\textbf{Effect of angular margin contrastive learning.} To better understand the effect of our angular margin contrastive method, we show in Table \ref{tab:similarity_score_video_text_pairs} examples of videos and language captions, along with the similarity scores which are generated by the baseline CLIP-ViP model and the model trained with our proposed subtractive contrastive learning strategy. We observe that when the caption consists of less details, \textit{i.e.} it aligns more weakly with the video content, our model assigns a lower similarity score compared to CLIP-ViP. In contrast, when the language caption becomes more detailed, we assign a higher score while CLIP-ViP only slightly elevates the score. These examples demonstrate that our angular margin contrastive strategy can control the similarity level for weakly aligned video-text pairs while being able to adapt to cases of strongly aligned pairs. 

\noindent\textbf{Effect of MLP-parameterized weighting function.} To closely study the impact of our MLP-based weighting function, we show the relative R@1 and the relative accuracy improvement of the text-video retrieval tasks in Figure \ref{fig:improvement_figure}a and \ref{fig:improvement_figure}b, respectively. Both figures demonstrate that we accomplish higher level of performance improvement on samples that belong to minority groups, \textit{i.e.} unpopular topic groups and answer labels. Such results intuitively substantiate the productivity of our MLP-based weighting function and showcase its capacity to learn the effective strategy to control the training effect of the training samples.

%% file: files/05_conclusion.tex
\vspace{-8mm}
\section{Conclusion}
\vspace{-2mm}
In this paper, we propose a meta-optimized contrastive framework to enhance video-language representation learning. In particular, we propose a contrastive learning framework with a subtractive margin between positive video and language to regularize their representations from reaching perfect similarity. Our framework also utilizes an MLP network that maps training losses to sample weights to enable the model to dynamically adjust the focus on data samples during training. Combined with a strategy to utilize large-vision language model to augment video-text data, our framework achieves superior results on commonly used video question answering and text-video retrieval tasks. Our framework is also applicable to a wide array of model architectures, which can promote its implementation in practical applications.

%% file: files/06_acknowledgement.tex
\vspace{5pt}
\noindent\textbf{Acknowledgement.} This research/project is supported by the National Research Foundation, Singapore under its AI Singapore Programme (AISG Award No: AISG3-PhD-2023-08-051T). Thong Nguyen is supported by a Google Ph.D. Fellowship in Natural Language Processing.

%% file: files/appendix.tex
\newpage
\appendix
\section*{Appendix}
\section{Proof of Theorem 1}
\label{app:proof_theorem_1}

\begin{customthm}{1.}
Let $\lambda_{i,j}$ denote the angle between the representation of two samples $i, j$, $\mathcal{L}^{v,t}_{\textup{angular}, i}$ and $\mathcal{L}^{v,t}_{\textup{contrastive}, i}$ denote the training objectives with and without the angular margin, respectively. Then, if $\lambda_{i,i} \leq \frac{\pi}{2}$, the following inequality holds:
\begin{equation}
\left|\frac{\partial \mathcal{L}_{\textup{angular},i}^{v,t}}{\partial \lambda_{i,i}}\right| \leq \left|\frac{\partial \mathcal{L}_{\textup{contrastive}, i}^{v,t}}{\partial \lambda_{i,i}}\right| 
\end{equation}
\end{customthm}

\begin{proof}
From the formulation of $\mathcal{L}_{\textup{angular}, i}^{v,t}$, we have the following equation:
\begin{equation}
\mathcal{L}_{\textup{angular},i}^{v,t} = -\frac{\cos\left([\lambda_{i,i}-\mu]_{+}\right)}{\tau} + \log \left(e^{\frac{\cos\left([\lambda_{i,i}-\mu)]_{+}\right)}{\tau}} + \sum\limits_{j \neq i} e^{\frac{\cos(\lambda_{i,j})}{\tau}}\right).
\end{equation}
Differentiating $\mathcal{L}_{\textup{angular},i}^{v,t}$ with respect to $\lambda_{i,i}$, we obtain:
\begin{align}
\frac{\partial \mathcal{L}_{\textup{angular},i}^{v,t}}{\partial \lambda_{i,i}} &= \frac{\sin\left([\lambda_{i,i}-\mu]_{+}\right)}{\tau} - \frac{\sin\left([\lambda_{i,i}-\mu]_{+}\right) \cdot e^{\frac{\cos(\lambda_{i,j})}{\tau}}}{e^{\frac{\cos\left([\lambda_{i,i}-\mu)]_{+}\right)}{\tau}} + \sum\limits_{j \neq i} e^{\frac{\cos(\lambda_{i,j})}{\tau}}} \\
&= \frac{\frac{\sin\left([\lambda_{i,i}-\mu]_{+}\right)}{\tau} \cdot \sum\limits_{j \neq i} e^{\frac{\cos(\lambda_{i,j})}{\tau}}}{e^{\frac{\cos\left([\lambda_{i,i}-\mu)]_{+}\right)}{\tau}} + \sum\limits_{j \neq i} e^{\frac{\cos(\lambda_{i,j})}{\tau}}}
\end{align}
Analogously, we achieve the derivative of $\mathcal{L}_{\textup{contrastive}, i}^{v,t}$ with respect to $\lambda_{i,i}$:
\begin{equation}
\frac{\partial \mathcal{L}_{\textup{contrastive}, i}^{v,t}}{\partial\lambda_{i,i}} = \frac{\frac{\sin\left([\lambda_{i,i}]_{+}\right)}{\tau} \cdot \sum\limits_{j \neq i} e^{\frac{\cos(\lambda_{i,j})}{\tau}}}{e^{\frac{\cos\left([\lambda_{i,i})]_{+}\right)}{\tau}} + \sum\limits_{j \neq i} e^{\frac{\cos(\lambda_{i,j})}{\tau}}}
\end{equation}
Dividing the norm of $\frac{\partial \mathcal{L}_{\textup{angular},i}^{v,t}}{\partial \lambda_{i,i}} $ by the norm of $\frac{\partial \mathcal{L}_{\textup{contrastive}, i}^{v,t}}{\lambda_{i,i}}$ yields that:
\begin{equation}
\begin{split}
\left|\frac{\partial \mathcal{L}_{\textup{angular},i}^{v,t}}{\partial \lambda_{i,i}}\right| / \left|\frac{\partial \mathcal{L}_{\textup{contrastive}, i}^{v,t}}{\partial\lambda_{i,i}}\right| = \frac{\sin\left([\lambda_{i,i} - \mu]_{+}\right)}{\sin\left(\lambda_{i,i}\right)} \cdot \frac{e^{\frac{\cos\left([\lambda_{i,i})]_{+}\right)}{\tau}} + \sum\limits_{j \neq i} e^{\frac{\cos(\lambda_{i,j})}{\tau}}}{e^{\frac{\cos\left([\lambda_{i,i}-\mu)]_{+}\right)}{\tau}} + \sum\limits_{j \neq i} e^{\frac{\cos(\lambda_{i,j})}{\tau}}}.
\end{split}
\end{equation}
Since, $\lambda_{i,i} \leq \frac{\pi}{2}$, we have $\sin\left([\lambda_{i,i} - \mu]_{+}\right) \leq \sin(\lambda_{i,i})$ and $\cos(\lambda_{i,i}) \leq \cos\left([\lambda_{i,i} - \mu]_{+}\right)$. Therefore,
\begin{equation}
\left|\frac{\partial \mathcal{L}_{\textup{angular},i}^{v,t}}{\partial \lambda_{i,i}}\right| / \left|\frac{\partial \mathcal{L}_{\textup{contrastive}, i}^{v,t}}{\partial\lambda_{i,i}}\right| \leq 1.
\end{equation}
This concludes our proof.  \qed
\end{proof}

\section{Derivation of the Update Rule in Theoretical Analysis}
\label{app:derivation_detail_update_rule_mlp}
\begin{proof}
Recalling Eq. (\ref{eq:update_mlp_network}), we have the update rule as follows:
\begin{equation}
\theta^{(k+1)} = \theta^{(k)} - \frac{\beta}{M} \sum\limits_{i=1}^{M} \nabla_{\theta} \mathcal{L}_{i}^{\text{meta}} \left( \hat{\Theta}^{(k)}\right).
\end{equation}
We continue our derivation using chain rule:
\begin{equation}
\theta^{(k+1)} = \theta^{(k)} - \frac{\beta}{M} \sum\limits_{i=1}^{M} \frac{\partial \mathcal{L}_{i}^{\text{meta}} \left(\hat{\Theta}^{(k)}\right)}{\partial \hat{\Theta}^{(k)}}\frac{\partial \hat{\Theta}^{(k)}}{\partial\theta}.
\end{equation}
Taking into account the update rule for the parameters of our video-language model in Eq. (\ref{eq:update_video_language_model}), we have:
{\footnotesize
\begin{equation}
\hspace{-20pt}
\begin{split}
&\theta^{(k+1)} = \theta^{(k)} - \frac{\beta}{M} \sum\limits_{i=1}^{M} \frac{\partial \mathcal{L}_{i}^{\text{meta}} \left( \hat{\Theta}^{(k)}\right)}{\partial \hat{\Theta}^{(k)}} \cdot \left(-\frac{\partial\frac{\alpha}{B}\sum\limits_{j=1}^{B} w_{i} \left(\theta^{(k)},\mathcal{L}_{j}^{\text{train}}\left(\Theta^{(k)}\right)\right) \nabla_{\Theta} \mathcal{L}_{j}^{\text{train}}\left(\Theta^{(k)}\right)}{\partial \theta}\right) \\
&= \theta^{(k)} + \frac{\alpha\beta}{BM} \sum\limits_{i=1}^{M}\sum\limits_{j=1}^{B} \nabla_{\hat{\Theta}^{(k)}} \mathcal{L}_{i}^{\text{meta}} \left(\hat{\Theta}\right)\Big|_{\hat{\Theta}^{(k)}} \nabla_{\Theta} \mathcal{L}_{j}^{\text{train}}\left(\Theta\right)\Big|_{\Theta^{(k)}} \nabla_{\theta} w\left(\theta^{(k)}, \mathcal{L}_{j}^{\text{train}}\left(\Theta^{(k)}\right)\right)\Big|_{\theta^{(k)}}.
\end{split}
\end{equation}}
If we assign:
\begin{equation}
G_{ij} = \nabla_{\hat{\Theta}^{(k)}} \mathcal{L}_{i}^{\text{meta}} \left( \hat{\Theta}\right)\Big|_{\hat{\Theta}^{(k)}} \nabla_{\Theta} \mathcal{L}_{j}^{\text{train}}\left(\Theta\right)\Big|_{\Theta^{(k)}},
\end{equation}
then we obtain the target result:
\begin{equation}
\theta^{(k+1)} = \theta^{(k)} + \frac{\alpha\beta}{BM} \sum_{i=1}^{B}\sum_{j=1}^{M} G_{ij} \frac{\partial w\left(\theta^{(k)}, \mathcal{L}_{j}^{\text{train}}\left(\Theta^{(k)}\right)\right)}{\partial\theta}\Big|_{\theta^{(k)}}
\end{equation}
\end{proof}

\section{Datasets}
\label{app:datasets}
\noindent\textbf{MSRVTT} \citep{xu2016msr} is composed of 10K YouTube videos, each of which possesses 20 sentence-level descriptions. For text-video retrieval, we follow the standard protocol to train on 9K videos and evaluate on the 1K-A testing split. For videoQA, MSRVTT is formatted into 243K open-ended questions. We adopt a split of 149K/12K/73K train/val/test split to evaluate our framework.

\noindent\textbf{MSVD} \citep{chen2011collecting} comprises 47K open-ended questions over 2K videos. We employ a ratio of 30K/6K/13K to split the questions into training, validation, and testing sets, respectively. 

\noindent\textbf{TGIF-QA-R} \citep{jang2017tgif, peng2021progressive} assesses the spatial-temporal reasoning ability to answer questions about GIF videos. There are three subsets in the dataset: (i) Action: to identify the repeated action; (ii) Transition: to recognize the action before or after some event; (iii) Frames: to answer questions regarding a particular frame in the video. TGIF-Action and TGIF-Transition support multi-choice videoQA, and TGIF-Frame supports open-ended videoQA. Peng et al. (2021) \citep{peng2021progressive} discover that TGIF-QA exhibits an answer bias that enables a model to determine the answer without relating to the video and question. They construct a debiased version of TGIF-QA, \textit{i.e.} TGIF-QA-R, in which they gather answer choices and randomly re-draw the distractor options for each question. To prevent the bias from impact the evaluation of our learned video-language representations, we employ the TGIF-QA-R variant in our experiments.

\noindent\textbf{NExT-QA} \citep{xiao2021next} specifically judges the causal and temporal reasoning. Its videos are inherited from the VidOR dataset \citep{shang2019annotating}, and used to construct three question categories, \textit{i.e.} Causal, Descriptive, and Temporal. NExT-QA supports the multi-choice videoQA setting, where each question consists of five options with one correct answer and four distractor answers.

\noindent\textbf{Causal-VidQA} \citep{li2022representation} focuses upon visual commonsense reasoning and designs four types of questions, \textit{i.e.} Counterfactual, Descriptive, Explanatory, and Predictive. Analogous to NExT-QA, it mainly supports multi-choice videoQA which tasks a model to select the correct answer among five available options. 

\noindent \textbf{DiDeMo} \citep{anne2017localizing} is composed of 10K Flickr videos. Theses videos are manually annotated to result in 40K sentences. We follow common practice \citep{lei2022revealing, lei2021less, li2023lavender} to concatenate all annotated sentences of a video and evaluate paragraph-to-video and video-to-paragraph retrieval tasks.

\noindent \textbf{ActivityNet} \citep{activityNet} consists of 20K YouTube videos annotated with 100K captions. Similar to DiDeMo, we also concatenate captions of a video to conduct paragraph-to-video and video-to-paragraph retrieval experiments. We train our model and prior baselines on 10K videos, validate on 5K, and test on the remaining 5K videos.

\section{Implementation Details}
\label{app:implementation_details}
We test our learned video-language representations on video question answering (videoQA) and text$\leftrightarrow$video retrieval tasks.

\noindent\textbf{Video question answering (VideoQA).} For videoQA, with respect to the VIOLET-based model, to fairly compare with previous works \citep{fu2021violet, fu2023empirical}, we sparsely sample $N_V = 5$ video frames and utilize frame size of $224 \times 224$. We also adopt a max length of 30 words for every language question. We then use Video Swin Transformer \citep{liu2021video} with VideoSwin-Base pretrained on Kinetics-600 to encode the frame sequence, and pretrained BERT-Base embeddings to encode the language question and the answer. For the VGT-based model \citep{xiao2023contrastive}, we randomly sample $N_V = 32$ frames given a video input. We utilize Faster-RCNN \citep{ren2015faster} with the ResNet-101 backbone \citep{he_CVPR2016_resnet} pretrained on the Visual Genome dataset to extract visual frame features. For textual features, we employ the pretrained BERT-Base embedding layer. 

\noindent\textbf{Text$\leftrightarrow$video retrieval.} For text$\leftrightarrow$video retrieval, given a video, we randomly sample $N_V = 12$ frames and resize them to $224 \times 224$ images. We split a language query into word tokens with a maximum length of 70. We use CLIP visual encoder and CLIP textual encoder to encode video and language query, respectively.

\section{More examples of similarity scores for video-text pairs}
\label{app:example_similarity_score_video_text_pairs}
\begin{table}[h!]
\centering
\caption{Case study of similarity scores for video-text pairs.}
\label{app:tab_similarity_score_video_text_pairs}
\resizebox{\linewidth}{!}{
\begin{tabular}{c|p{0.5\linewidth}|c|c}
\hline
\textbf{Video} & \multicolumn{1}{c|}{\textbf{Caption}} & \textbf{CLIP-ViP score} & \textbf{Our score} \\ \hline
\multirow{6}{*}{\includegraphics[width=0.1\linewidth]{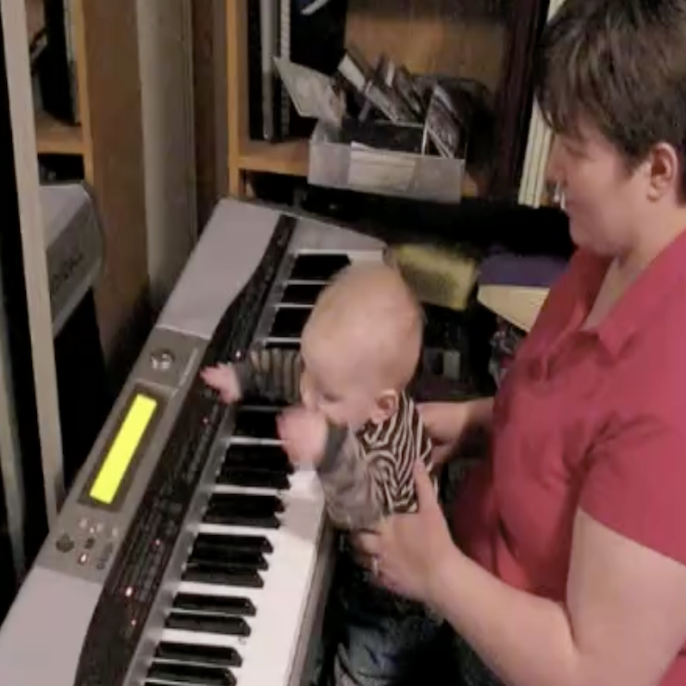} \includegraphics[width=0.1\linewidth]{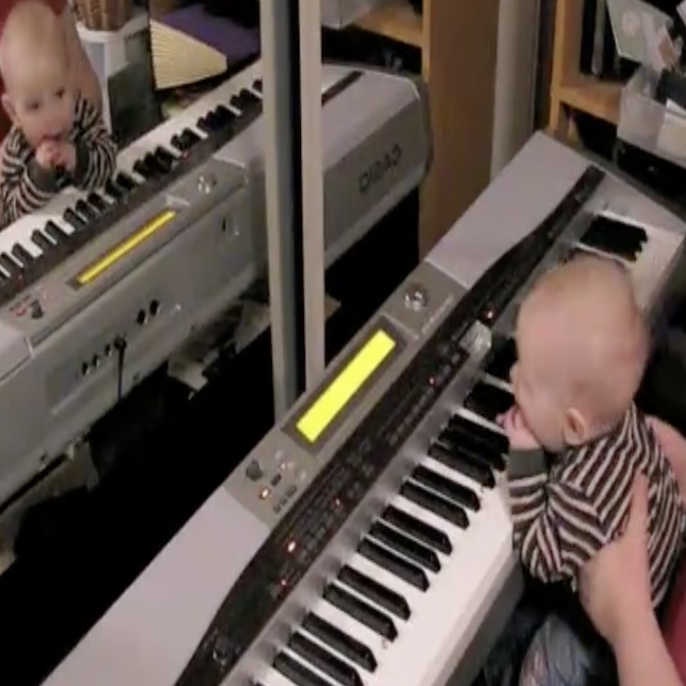}
\includegraphics[width=0.1\linewidth]{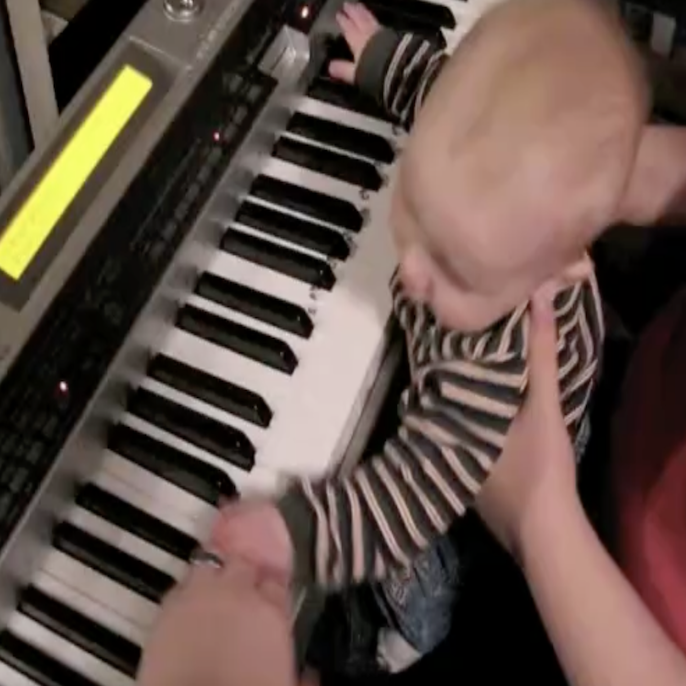} \includegraphics[width=0.1\linewidth]{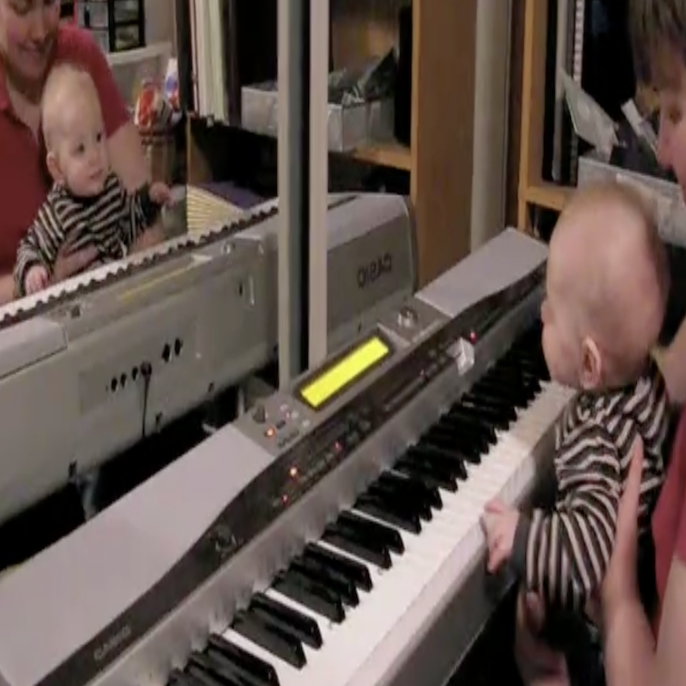}}                  &      a baby playing with a keyboard.                            &    0.3297                                         &            0.3277                           \\ \cline{2-4}
                                   &      the cameraman's hand moves into the screen to play keys and grab the babies hand. camera pans down to baby the cameraman reaches down and plays some keys on the piano.                               &   0.3580                                         &  0.3843                                      \\ \hline
\multirow{5}{*}{\includegraphics[width=0.1\linewidth]{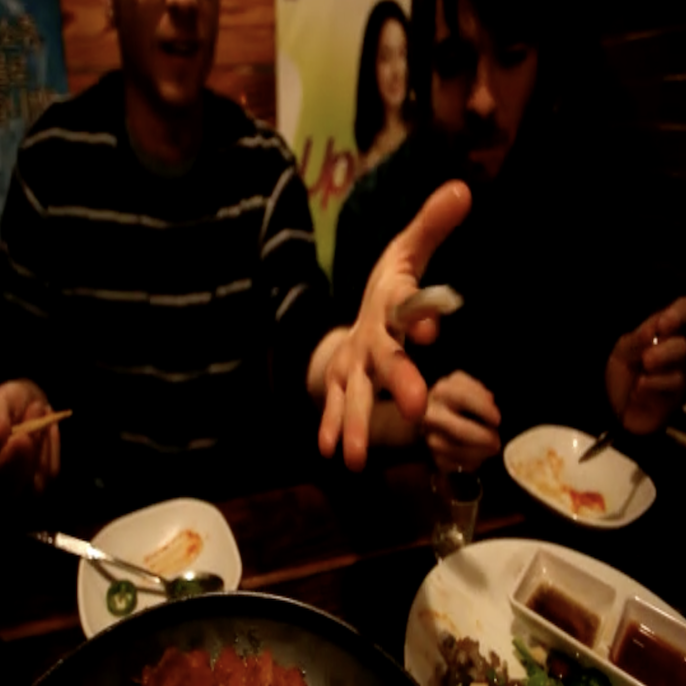} \includegraphics[width=0.1\linewidth]{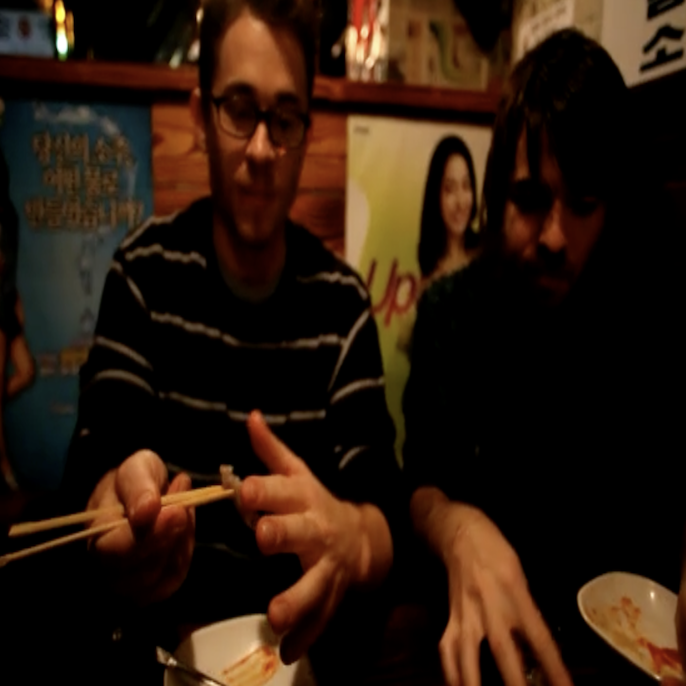}
\includegraphics[width=0.1\linewidth]{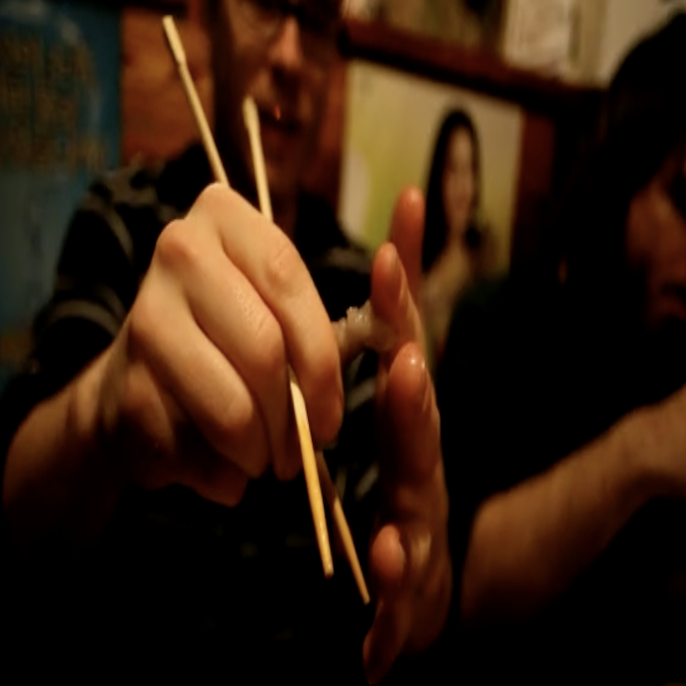} \includegraphics[width=0.1\linewidth]{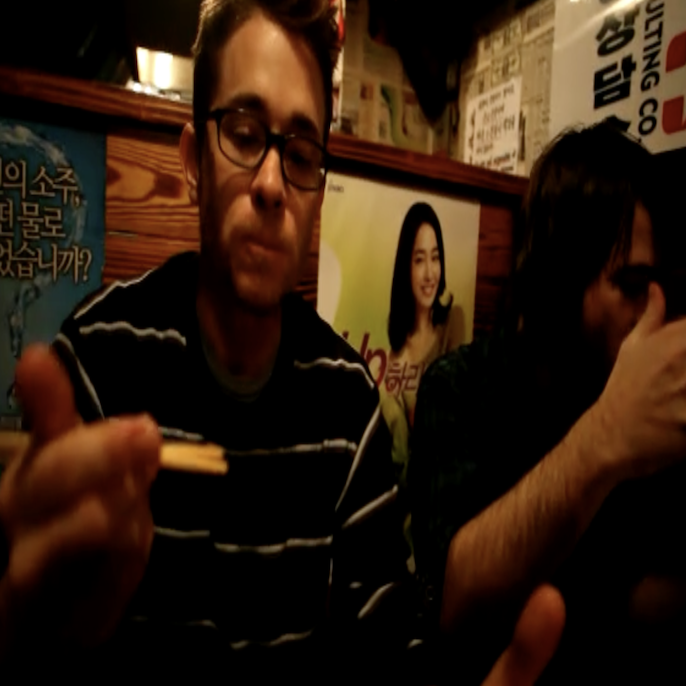}}                  &              a person in a striped shirt is holding a spoon.            &       0.2123                                      &           0.1949                        \\ \cline{2-4}
                                   &     a man takes a bite of food. Man eats thing on chopstick. Man pulls item off finger and eats it.                                &   0.3385                                         &   0.3504  \\    \hline
\multirow{5}{*}{\includegraphics[width=0.1\linewidth]{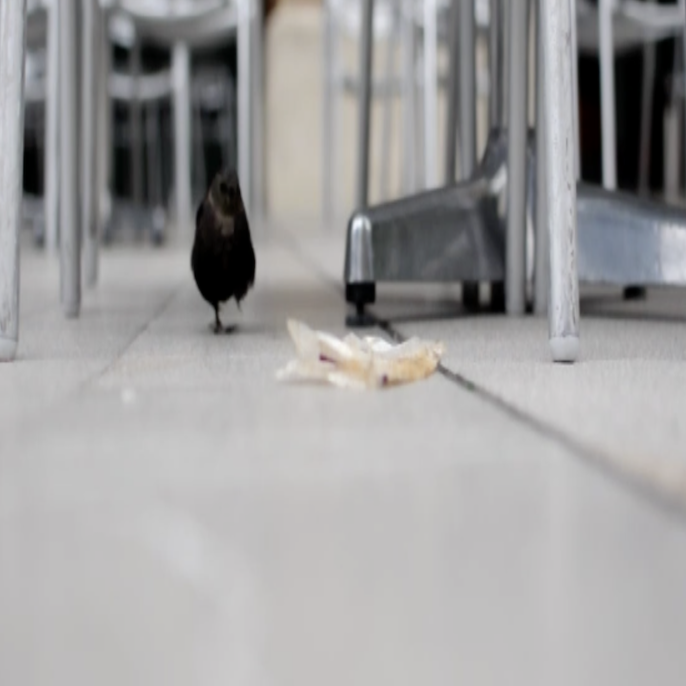} \includegraphics[width=0.1\linewidth]{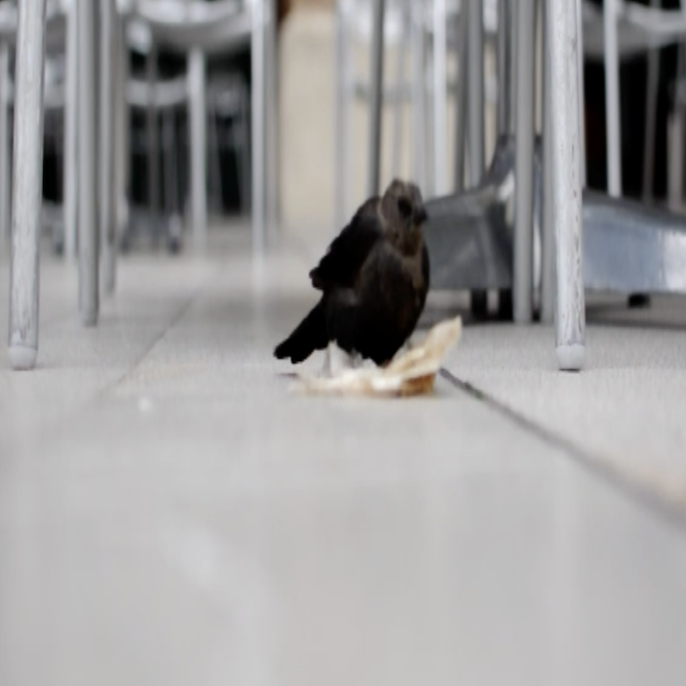}
\includegraphics[width=0.1\linewidth]{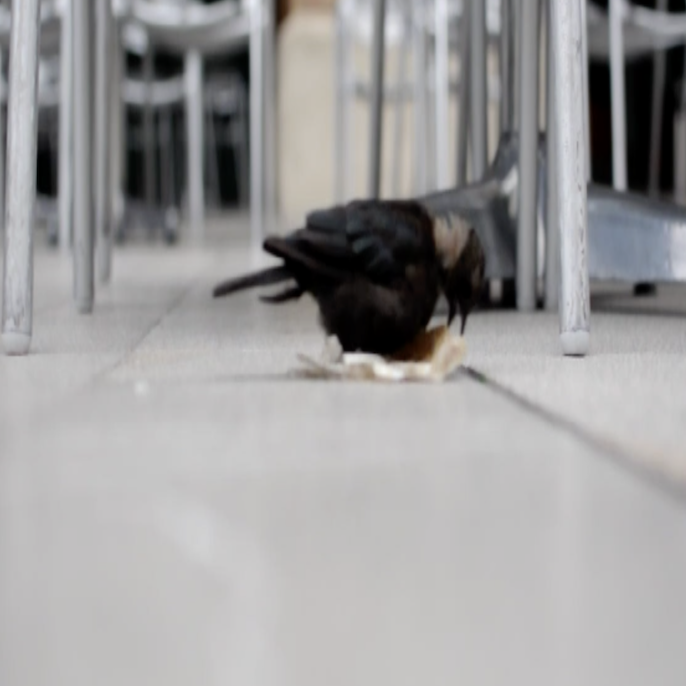} \includegraphics[width=0.1\linewidth]{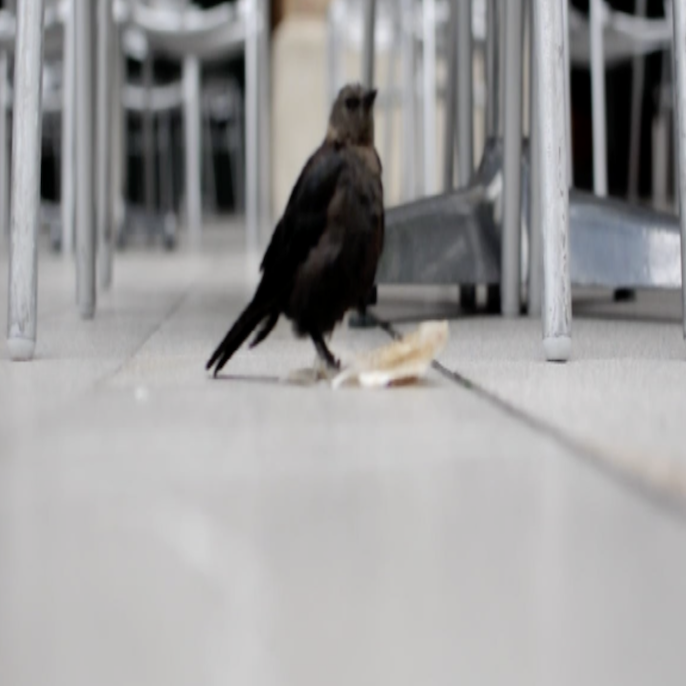}}                  &              a bird is sitting on the ground in the middle of a row of doors.           &       0.2470                                     &           0.2175                      \\ \cline{2-4}
                                   &     bird picks up a bag bird folds paper over. the bird lifts the napkin with its head. the bird begins to eat the bread                               &   0.3070                                         &   0.3242  \\    \hline
\multirow{4}{*}{\includegraphics[width=0.1\linewidth]{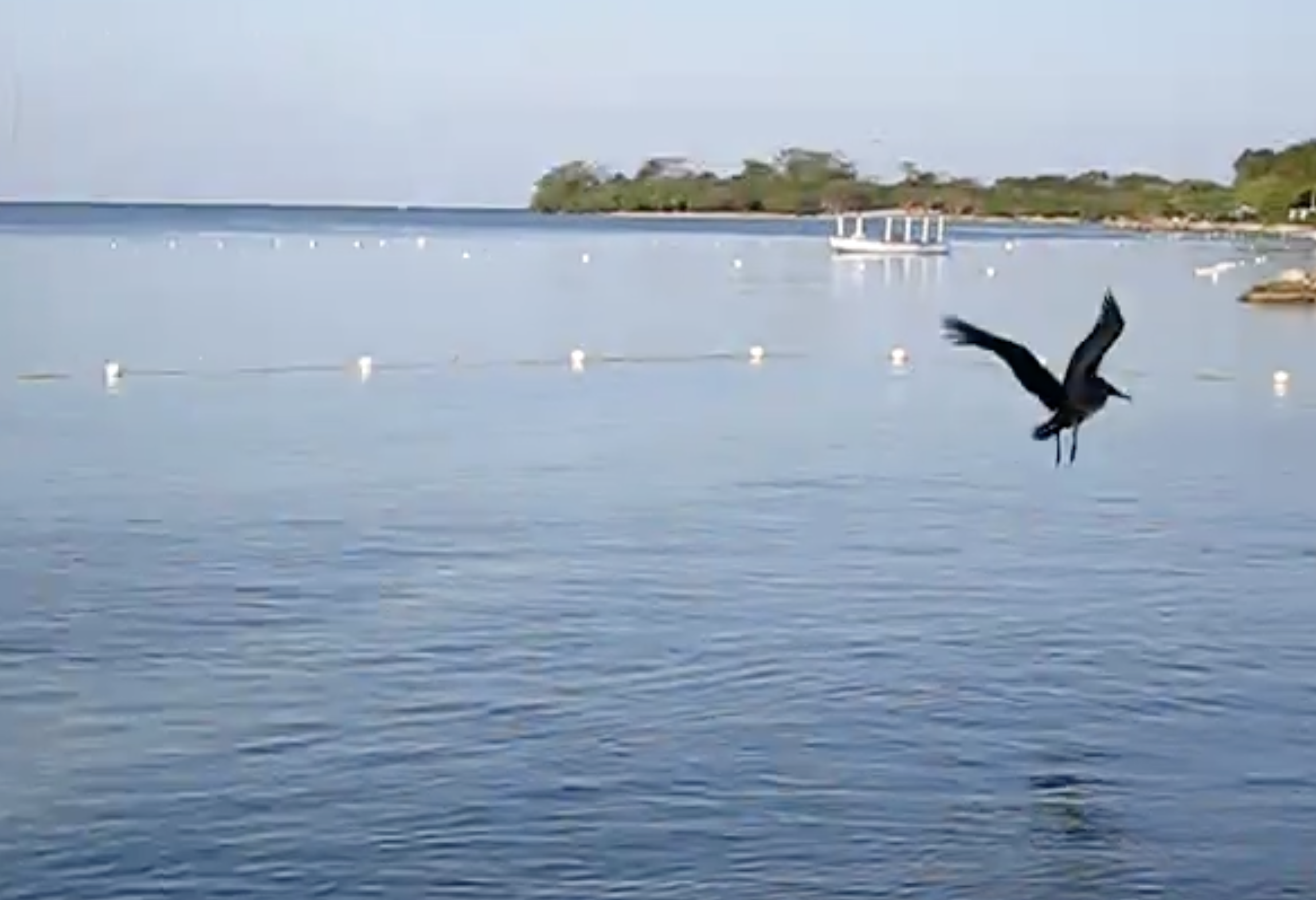} \includegraphics[width=0.1\linewidth]{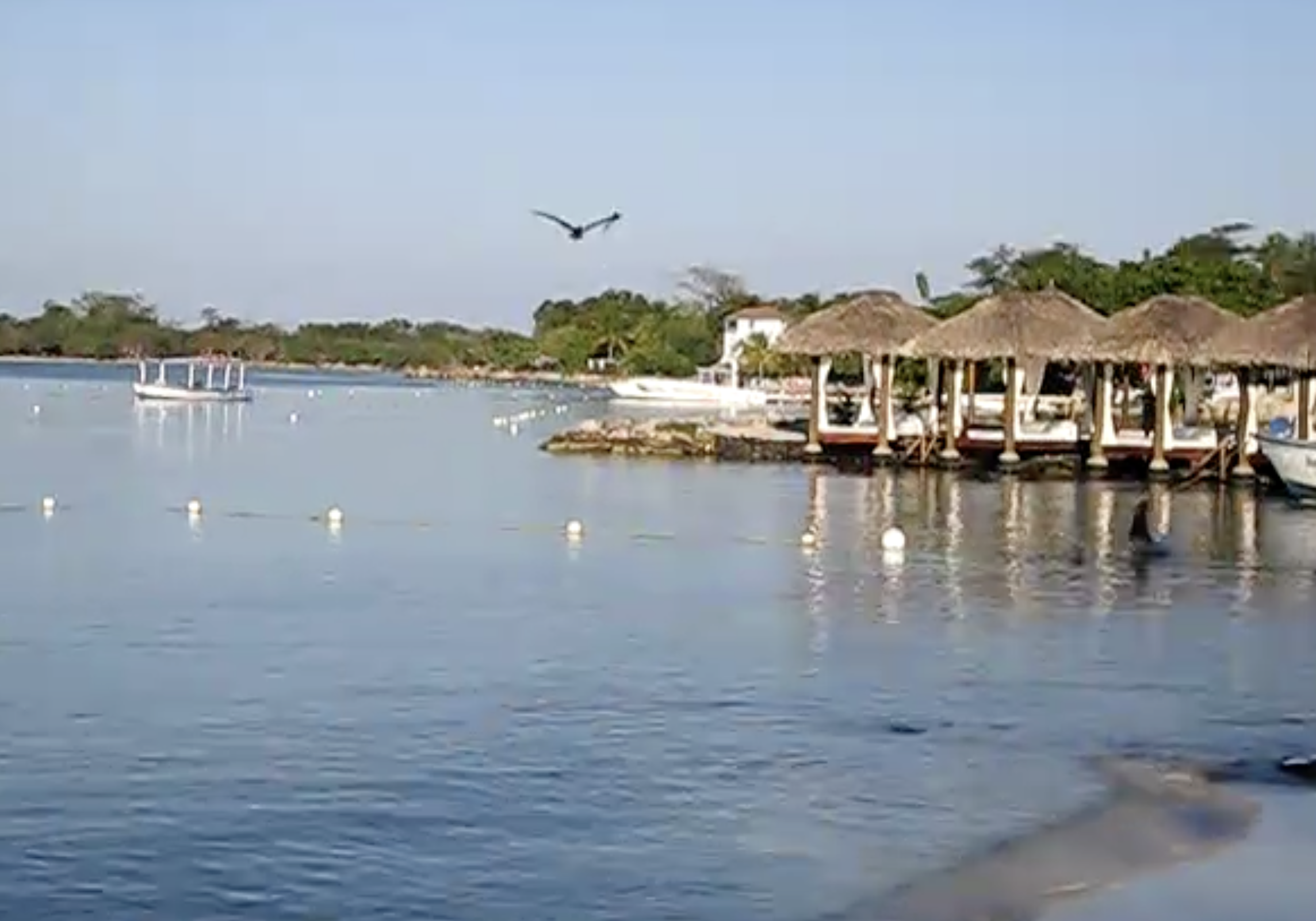}
\includegraphics[width=0.1\linewidth]{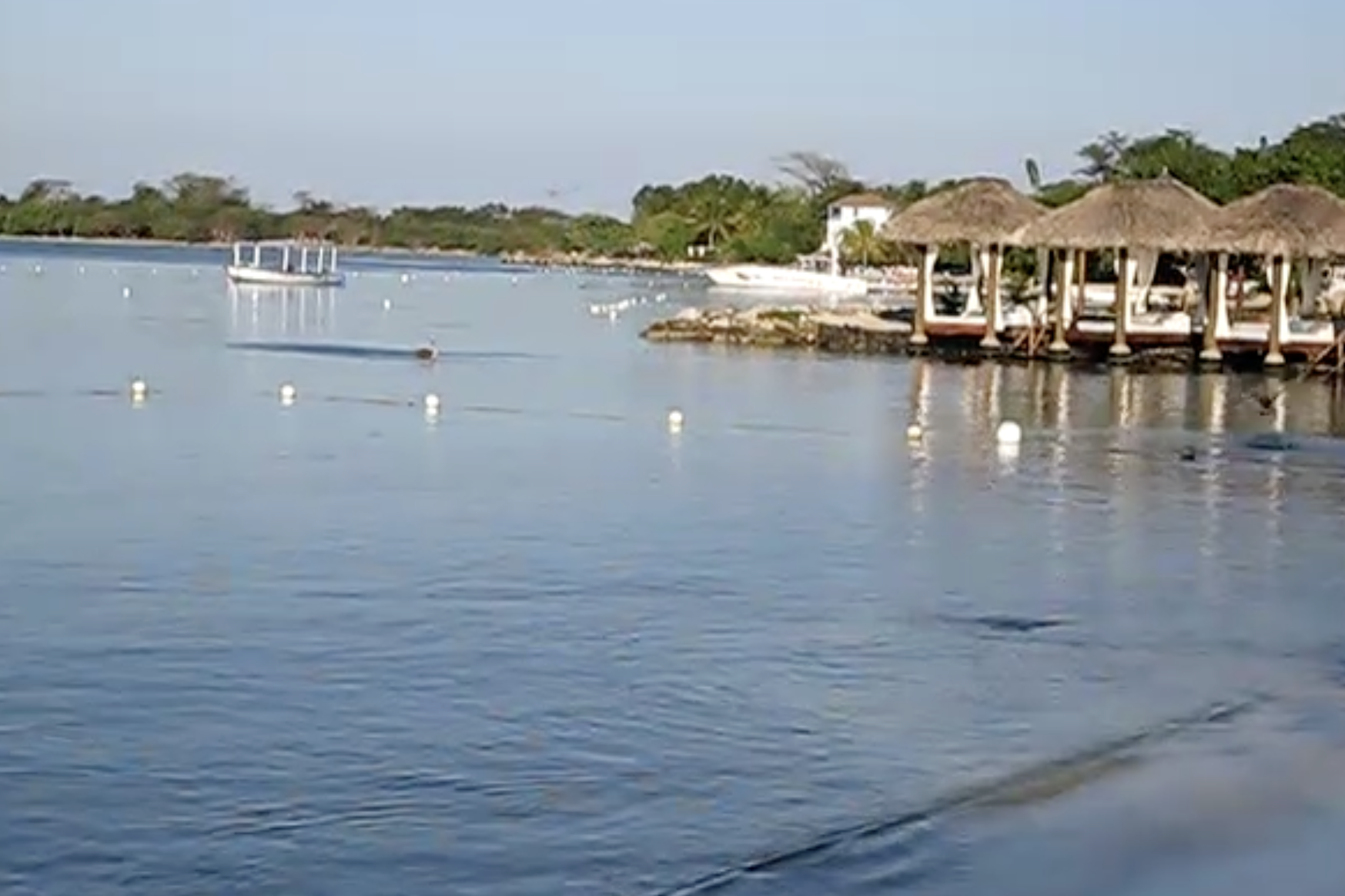} \includegraphics[width=0.1\linewidth]{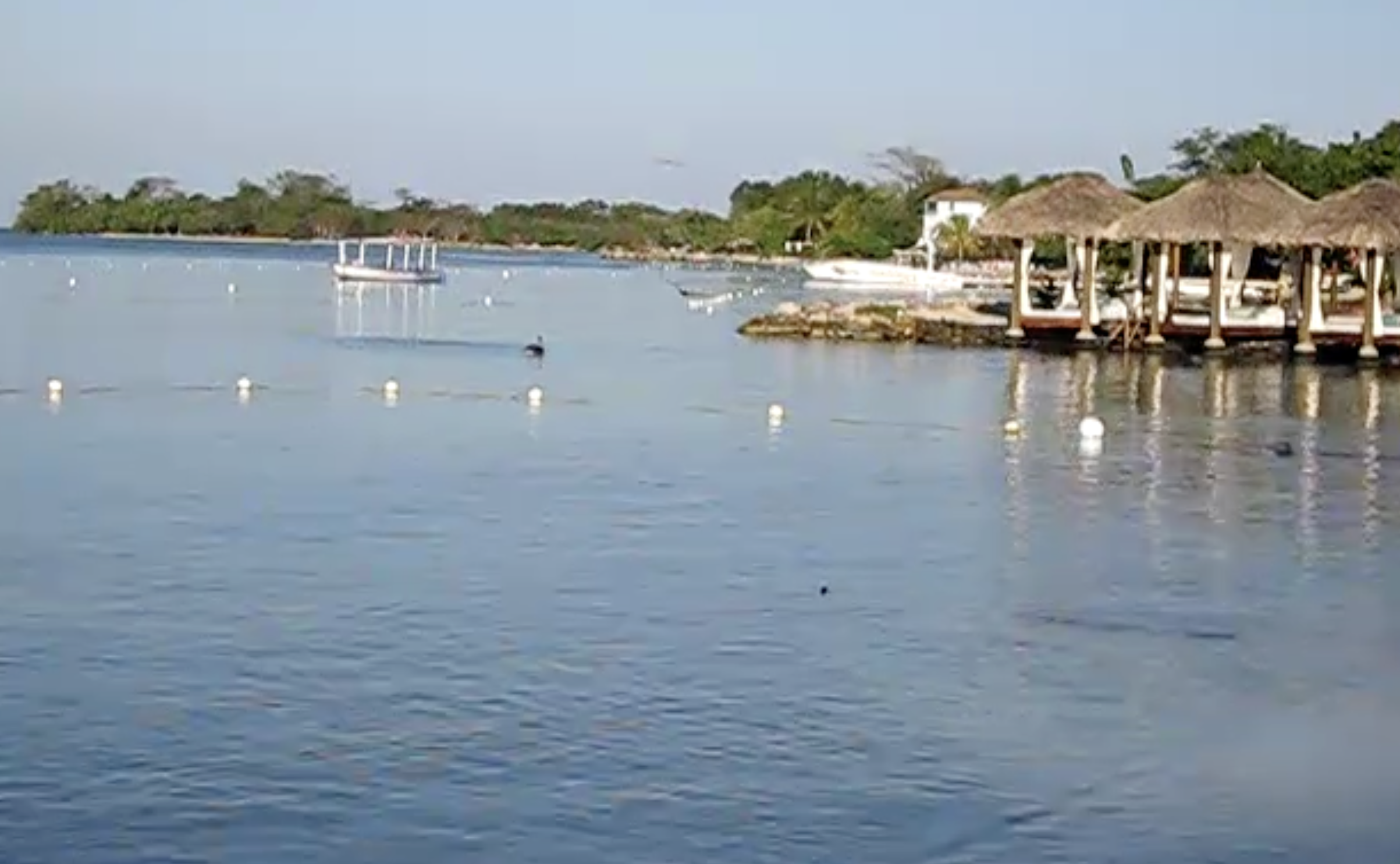}}                  &              a bird is flying above the water.            &       0.1991                                     &          0.1216                        \\ \cline{2-4}
                                   &     bird makes a few splashes in the water before flying off we see a bird skim the water ' fly away two birds fly away                              &   0.2771                                         &   0.3013  \\    \hline
\multirow{4}{*}{\includegraphics[width=0.1\linewidth]{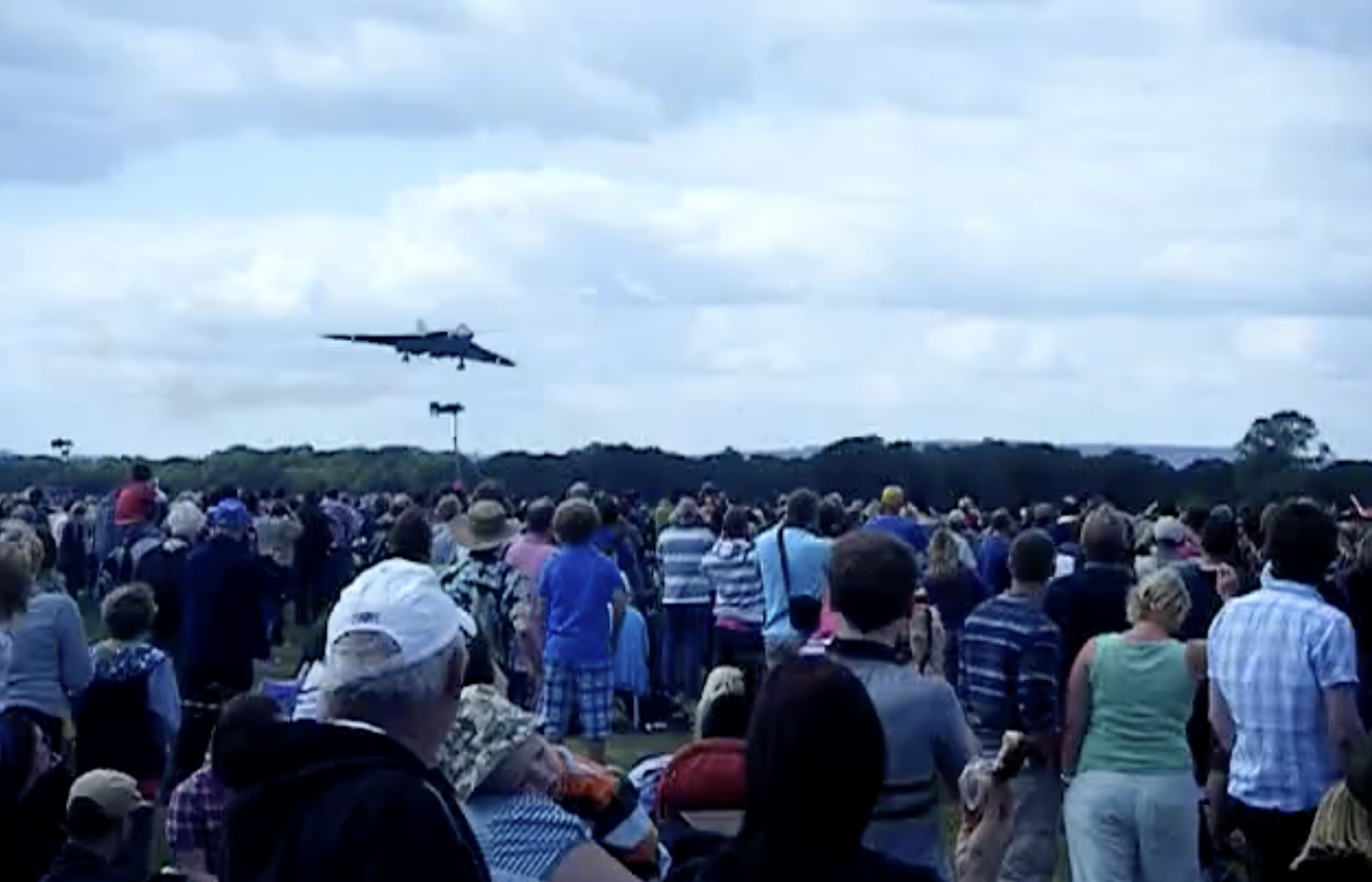} \includegraphics[width=0.1\linewidth]{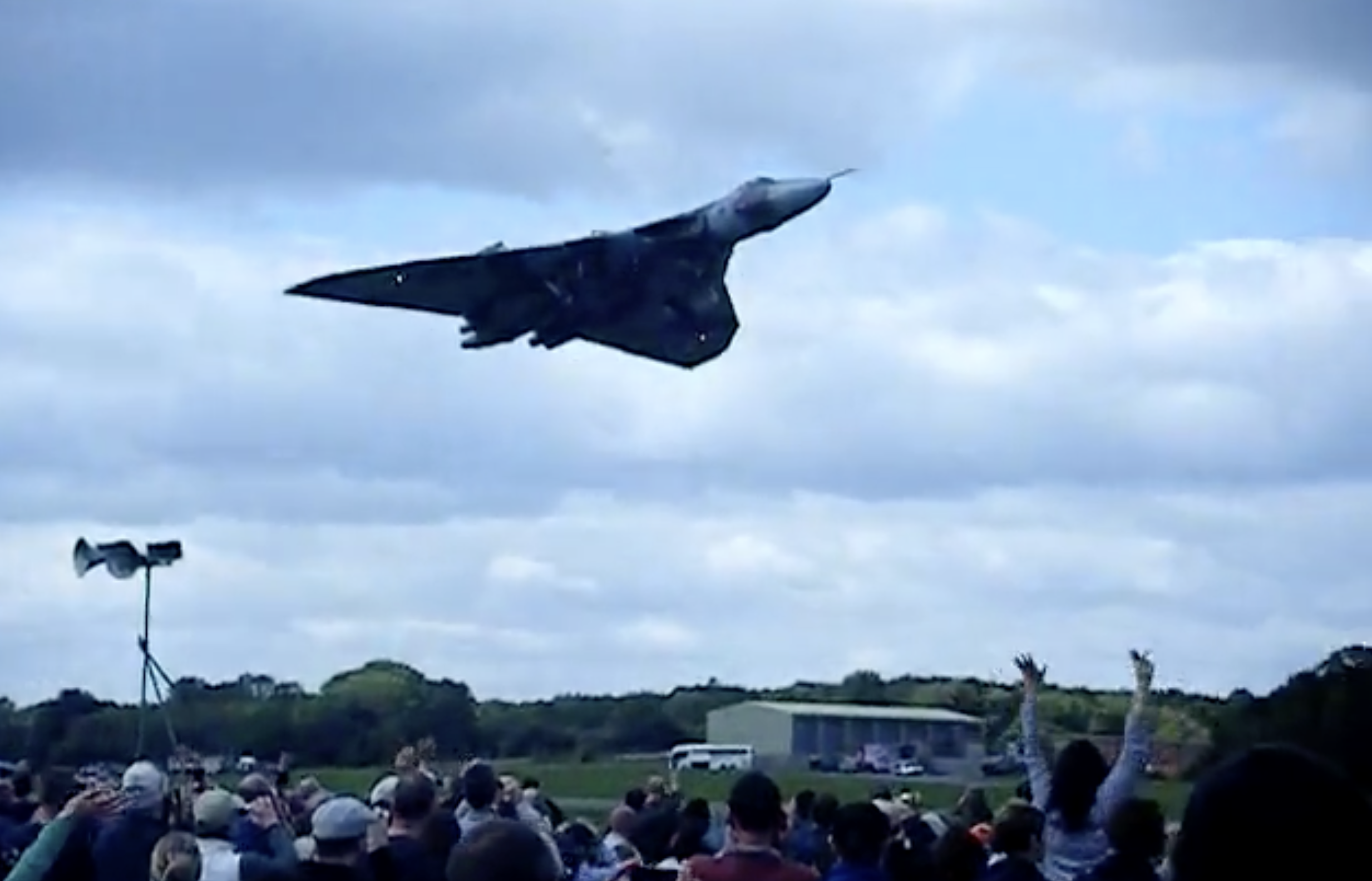}
\includegraphics[width=0.1\linewidth]{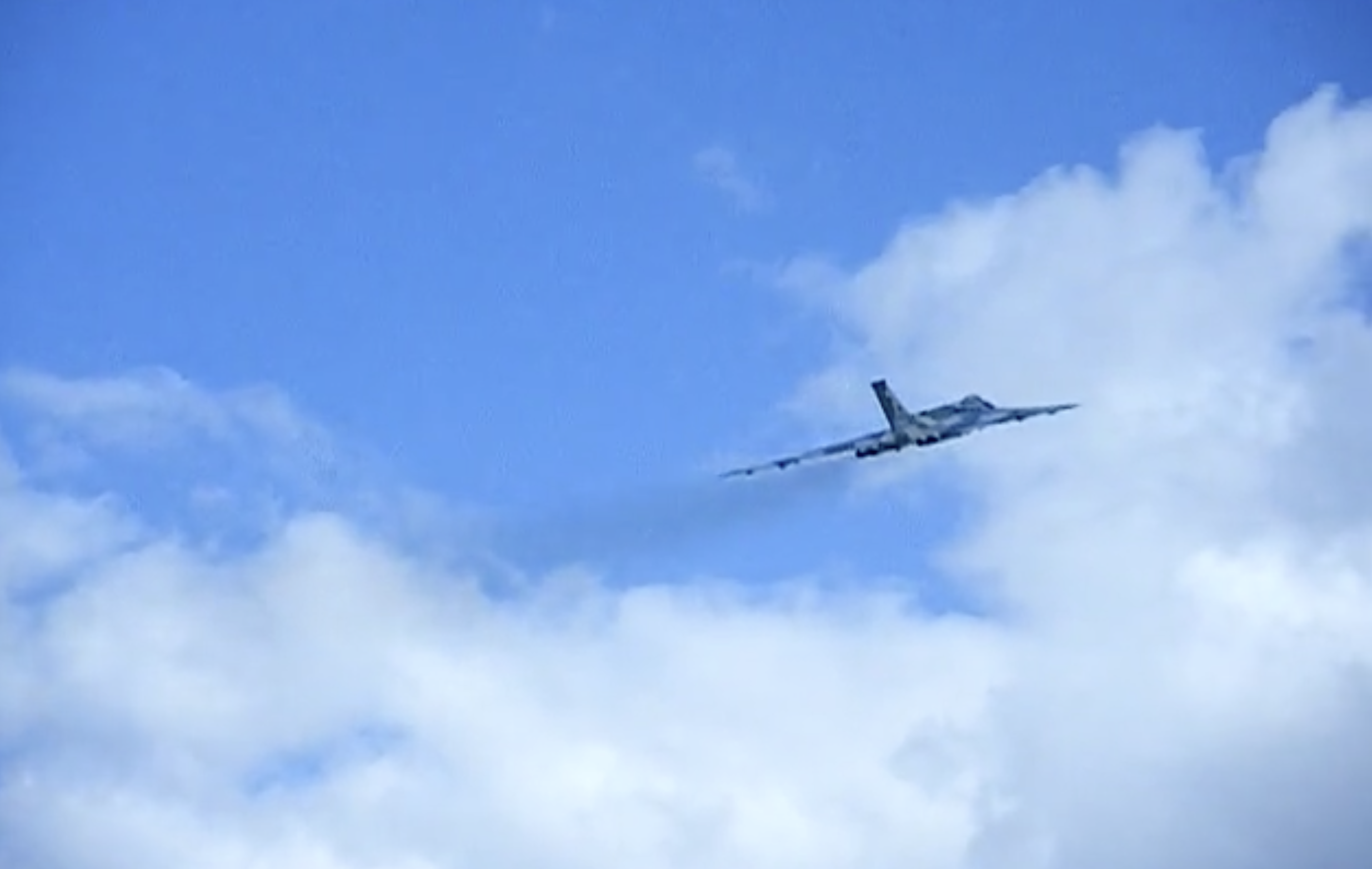} \includegraphics[width=0.1\linewidth]{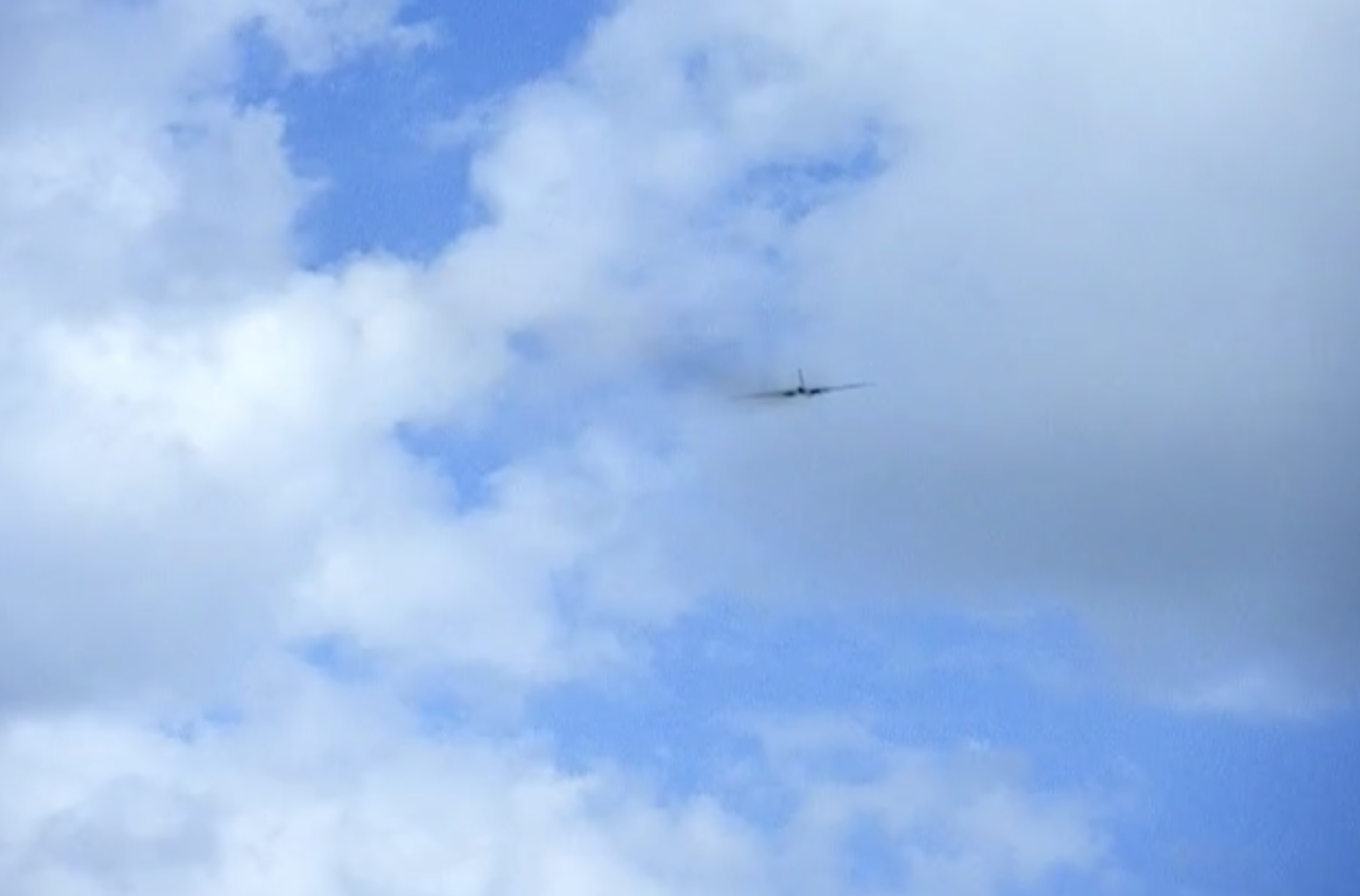}}                  &              a plane flying through the sky.            &       0.2331                                      &           0.2188                       \\ \cline{2-4}
                                   &     jet approaches the crowd then takes off again the plane is flying over the crowd. no crowd in these frames                               &   0.3503                                        &   0.3723  \\    \hline
\multirow{7}{*}{\includegraphics[width=0.1\linewidth]{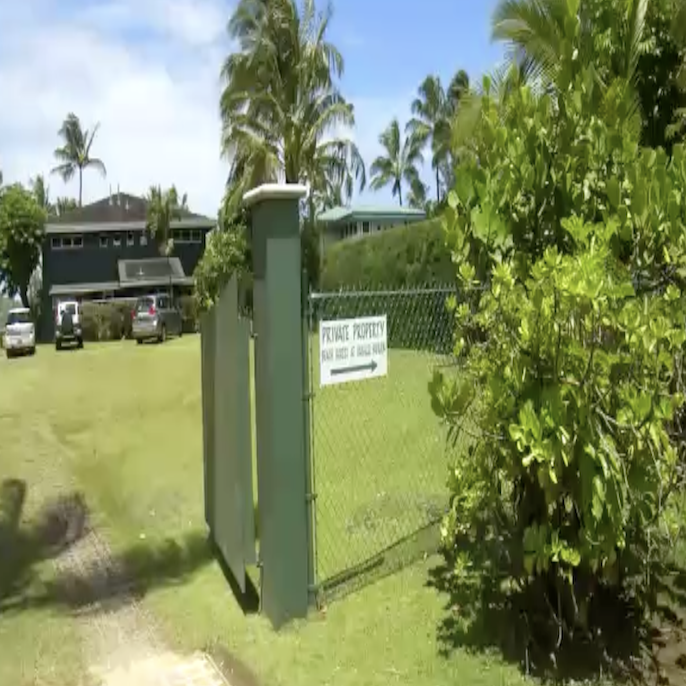} \includegraphics[width=0.1\linewidth]{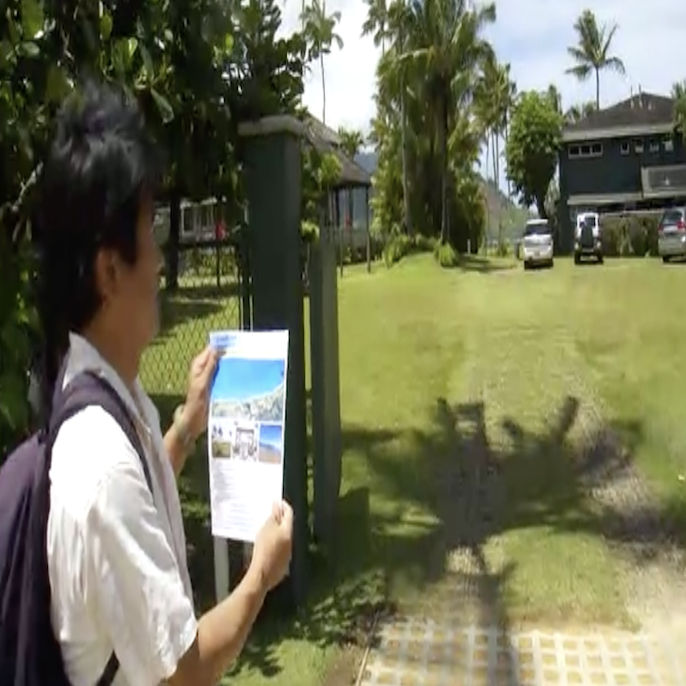}
\includegraphics[width=0.1\linewidth]{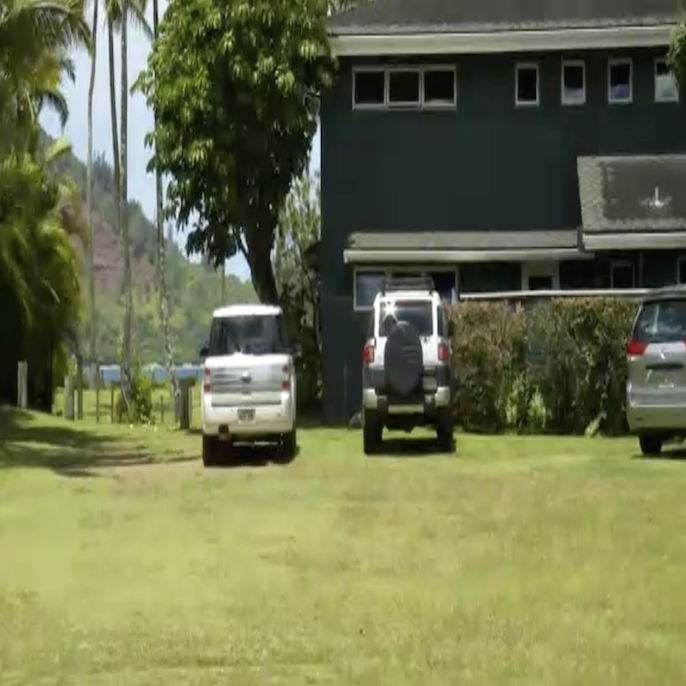} \includegraphics[width=0.1\linewidth]{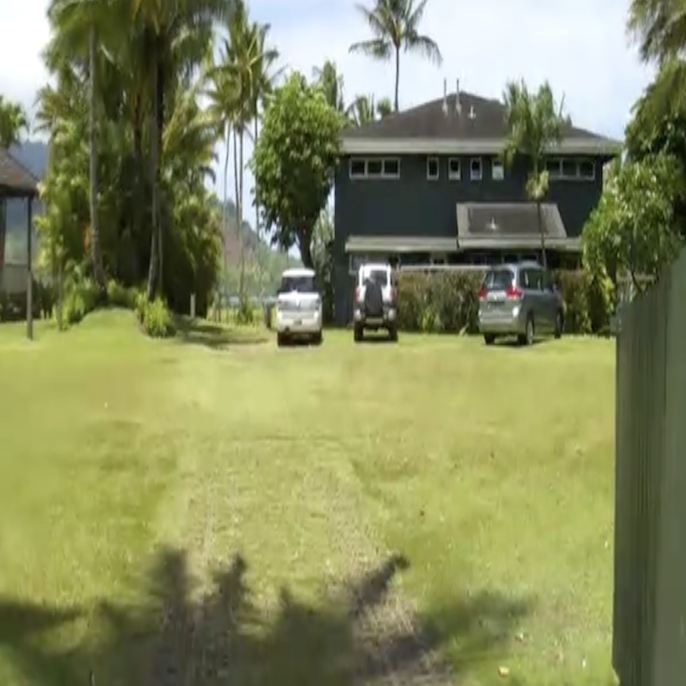}}                  &              a man is standing in front of a building with a sign on it.            &       0.1849                                      &           0.1713                       \\ \cline{2-4}
                                   &     guy enters into view camera pans left to show man with a map a person is standing at the gates of a location. a man holding up a pamphlet in front of a building. the person can be seen holding something                              &   0.3082                                         &   0.3153  \\    \hline
\end{tabular}}
\vspace{2mm}
\end{table}

In this appendix, we provide more examples of video-text pairs and their corresponding similarity scores computed by our model and the CLIP-ViP baseline in Table \ref{app:tab_similarity_score_video_text_pairs}.

\section{Additional Qualitative Results}
\label{app:more_qualitative_results}
In this appendix, we provide additional qualitative results of our proposed strategy which utilizes LVLM to augment video-text data.

\begin{figure*}[h!]
    \centering
    \caption{Examples of video input and our textual description.}
    \includegraphics[width=0.65\linewidth]{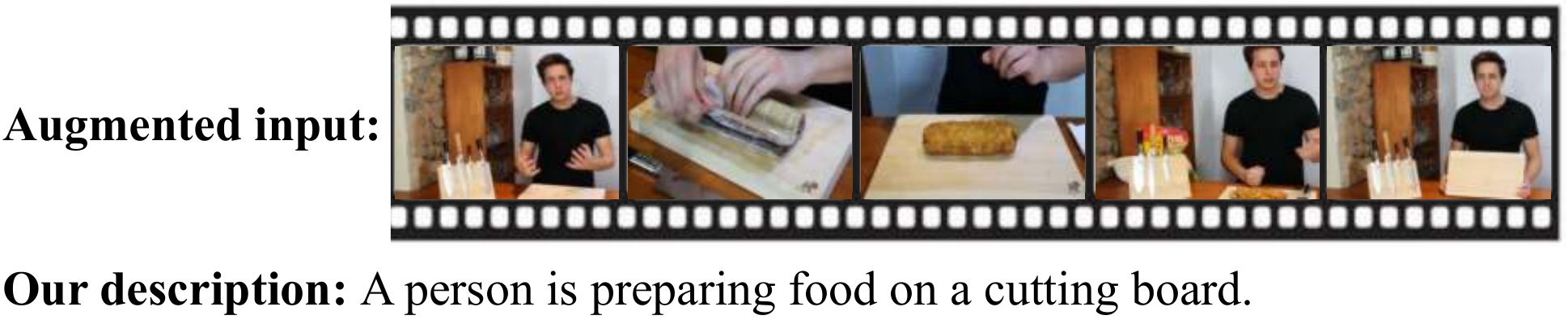}
\end{figure*}

\begin{figure*}[h!]
    \centering
    \includegraphics[width=0.65\linewidth]{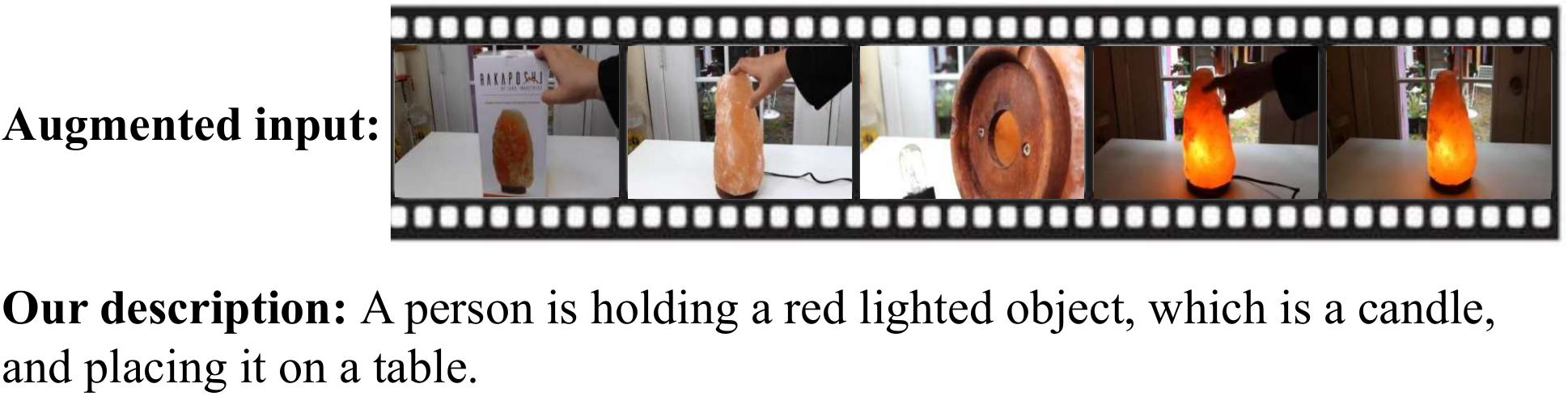}
\end{figure*}

\begin{figure*}[h!]
    \centering
    \includegraphics[width=0.65\linewidth]{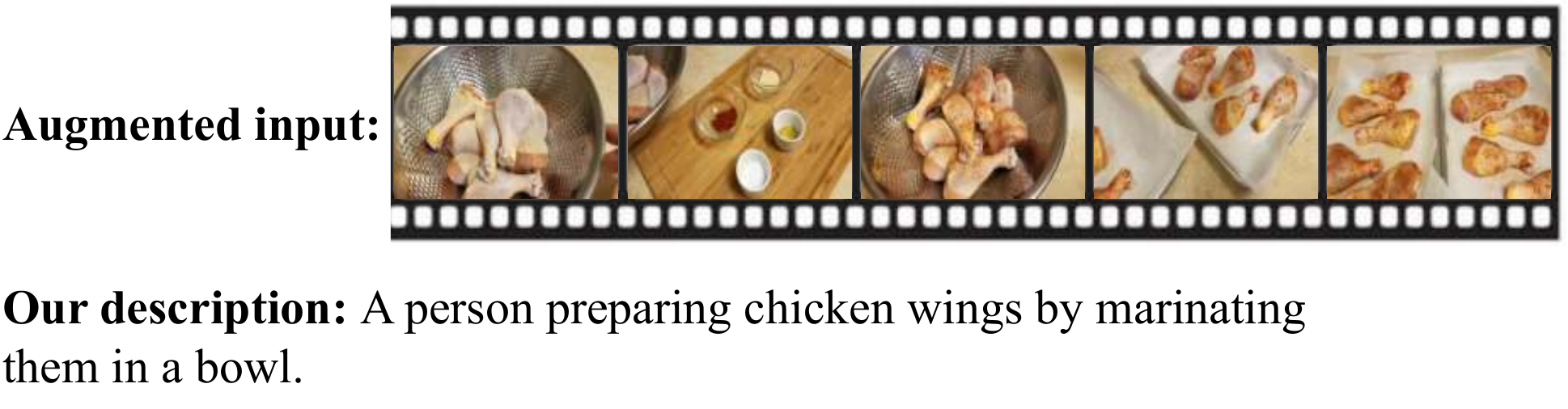}
\end{figure*}

\begin{figure*}[h!]
    \centering
    \includegraphics[width=0.65\linewidth]{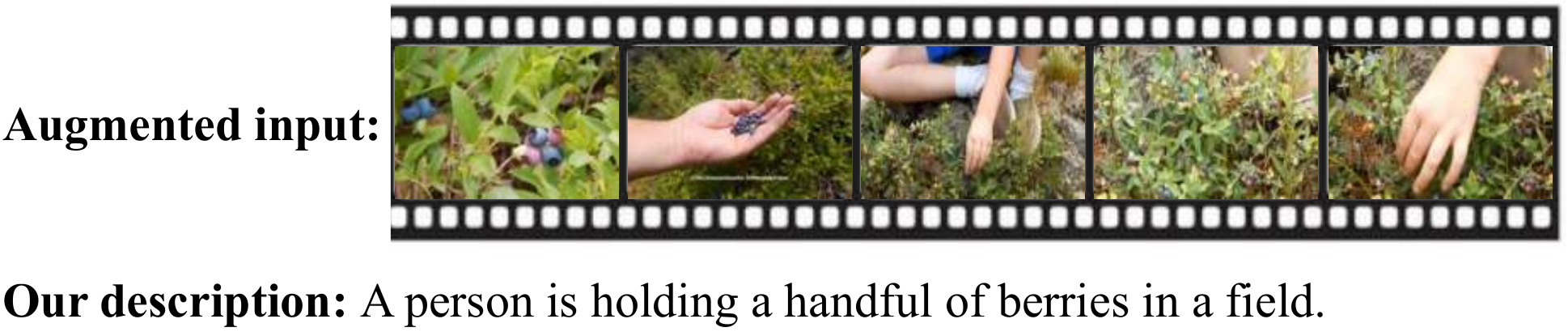}
\end{figure*}

\begin{figure*}[h!]
    \centering
    \includegraphics[width=0.65\linewidth]{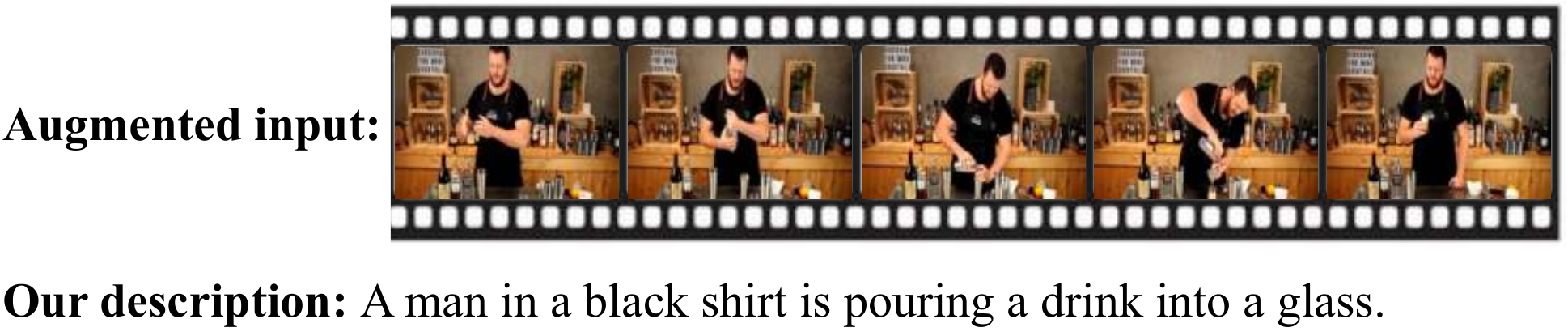}
\end{figure*}

\begin{figure*}[h!]
    \centering
    \includegraphics[width=0.65\linewidth]{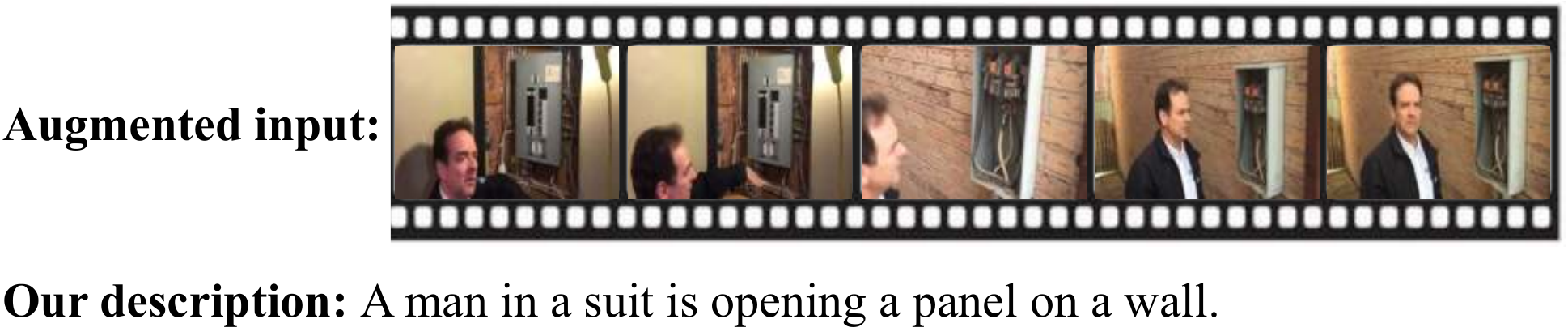}
\end{figure*}